\documentclass{article}

    \PassOptionsToPackage{numbers, compress}{natbib}
  \usepackage[preprint]{template-files/neurips_2024}

\usepackage[utf8]{inputenc} % allow utf-8 input
\usepackage[T1]{fontenc}    % use 8-bit T1 fonts
\usepackage{hyperref}       % hyperlinks
\usepackage{url}            % simple URL typesetting
\usepackage{booktabs}       % professional-quality tables
\usepackage{amsfonts}       % blackboard math symbols
\usepackage{nicefrac}       % compact symbols for 1/2, etc.
\usepackage{microtype}      % microtypography
\usepackage{xcolor}         % colors

\title{Expected Grad-CAM: Towards gradient faithfulness}

\author{
  Vincenzo Buono\\
  Halmstad University\\
  \texttt{vincenzo.buono@hh.se}
  \And
  Peyman S. Mashhadi \\
  Halmstad University\\
  \texttt{peyman.mashhadi@hh.se}
  \And
  Mahmoud Rahat\\
  Halmstad University\\
  \texttt{mahmoud.rahat@hh.se}
  \AND
  Prayag Tiwari\\
  Halmstad University\\
  \texttt{prayag.tiwari@hh.se}
  \And
  Stefan Byttner\\
  Halmstad University\\
  \texttt{stefan.byttner@hh.se} \\
}

\usepackage{import}
\usepackage{transparent}
\usepackage[acronym]{glossaries} % --> allowed? otherwise just use \def or \newcommand
\usepackage{multirow}
\usepackage{tikz}
\usepackage{soul}
\usepackage{adjustbox}

\usepackage{siunitx}
\usepackage{microtype}
\usepackage{graphicx}
\usepackage{subcaption}
\usepackage{booktabs} % for 

\usepackage{xspace}
\makeatletter
\DeclareRobustCommand\onedot{\futurelet\@let@token\@onedot}
\def\@onedot{\ifx\@let@token.\else.\null\fi\xspace}

\def\eg{\emph{e.g}\onedot} 
\def\ie{\emph{i.e}\onedot}

\def\wrt{w.r.t\onedot}

\makeatother

\usepackage{amsmath}
\usepackage{amssymb}
\usepackage{mathtools}
\usepackage{amsthm}
\theoremstyle{plain}
\newtheorem{theorem}{Theorem}[section]

\theoremstyle{definition}
\newtheorem{definition}[theorem]{Definition}

\theoremstyle{remark}
\newtheorem{remark}[theorem]{Remark}

\newtheoremstyle{lemma}
  {3pt}% Space above
  {3pt}% Space below
  {}% Body font
  {\parindent}% Indent amount
  {\bfseries}% Theorem head font
  {}% Punctuation after theorem head
  {.5em}% Space after theorem head
  {#1 3.#2}% Custom head spec
\newtheoremstyle{plemma}
  {3pt}% Space above
  {3pt}% Space below
  {}% Body font
  {\parindent}% Indent amount
  {\bfseries}% Theorem head font
  {}% Punctuation after theorem head
  {0em}% Space after theorem head
  {#1 3.#2}% Custom head spec

\theoremstyle{lemma}
\newtheorem{mdef}{Definition}
\theoremstyle{plemma}
\newtheorem{mprop}{Proposition}

  {\begin{mdef}[#2] \ifx\relax#2\relax\else(#2)\textbf{.}\fi \label{#1}\itshape}
  {\end{mdef}}
  
  {\begin{mprop}[#2] \textbf{.} \label{#1}\itshape}
  {\end{mprop}}
  
  {\vspace{3pt}\textbf{#2} (#3)\textbf{.}\hspace{.5em}\itshape}
  {\itshape\vspace{3pt}}
  
  {\vspace{3pt}\noindent\textbf{Remark:}\hspace{.1em}}
  {\itshape\vspace{3pt}}
  
\newenvironment{bsect}[1]
  {\vspace{0pt}\noindent\textbf{#1.}\hspace{.5em}}
  {\itshape\vspace{0pt}}
\usepackage{tikz}
\usepackage{scalerel}
\usepackage{pict2e}
\usepackage{tkz-euclide}
\usetikzlibrary{calc}
\usetikzlibrary{patterns,arrows.meta}
\usetikzlibrary{shadows}
\usetikzlibrary{external}

\usepackage{pgfplots}
\pgfplotsset{compat=newest}
\usepgfplotslibrary{statistics}
\usepgfplotslibrary{fillbetween}
\usepackage{tikz-3dplot}
\usepackage{xcolor}

\usepackage[skins]{tcolorbox}
\usepackage{xifthen}
\usetikzlibrary{fadings}
\usetikzlibrary{shadows}

\newsavebox{\measure@tikzpicture}
\NewEnviron{scaletikzpicturetowidth}[1]{%
  \def\tikz@width{#1}%
  \begin{lrbox}{\measure@tikzpicture}%
  \BODY
  \end{lrbox}%
  \pgfmathparse{#1/\wd\measure@tikzpicture}%
  \BODY
}
\newcommand{\tins}[0]{\hspace*{1em}}
\newcommand{\rsmargin}{2cm}
\newcommand{\rumargin}{3cm}
\newcommand{\enlrearged}[1]{
\newgeometry{left=\rsmargin, right=\rsmargin, top=\rumargin, bottom=\rumargin}
#1
\restoregeometry
}

\usepackage{cleveref}
\begin{document}

\maketitle

% abstract ----------------------------------
\begin{abstract}
    Although input-gradients techniques have evolved to mitigate and tackle the challenges associated with gradients, modern gradient-weighted CAM approaches still rely on vanilla gradients, which are inherently susceptible to the saturation phenomena. Despite recent enhancements have incorporated counterfactual gradient strategies as a mitigating measure, these local explanation techniques still exhibit a lack of sensitivity to their baseline parameter. Our work proposes a gradient-weighted CAM augmentation that tackles both the saturation and sensitivity problem by reshaping the gradient computation, incorporating two well-established and provably approaches: Expected Gradients and kernel smoothing. By revisiting the original formulation as the smoothed expectation of the perturbed integrated gradients, one can concurrently construct more faithful, localized and robust explanations which minimize infidelity. Through fine modulation of the perturbation distribution it is possible to regulate the complexity characteristic of the explanation, selectively discriminating stable features. Our technique, \textit{Expected Grad-CAM}, differently from recent works, exclusively optimizes the gradient computation, purposefully designed as an enhanced substitute of the foundational Grad-CAM algorithm and any method built therefrom. Quantitative and qualitative evaluations have been conducted to assess the effectiveness of our method.\footnote{Implementation available at \url{https://github.com/espressoshock/pytorch-expected-gradcam}.}
\end{abstract}

% abstract ----------------------------------

% sections ----------------------------------
\section{Introduction}
\label{sec:introduction}
%%%%%%%%%%%%%%%%%%%%%%%%%%%%%%%%%%%%%%%%%%

In recent years, deep neural networks (DNNs) have consistently achieved remarkable performances across a rapidly growing spectrum of application domains. Yet, their efficacy is often coupled with a \textit{black-box} operational behavior, commonly lacking transparency and explainability \cite{Samek2017ExplainableModels, Adadi2018PeekingXAI}. Such challenges have catalyzed a shift towards the research and development of \textit{Explainable AI} (xAI) methodologies, aimed at obtaining a deeper understanding of the intrinsic mechanisms and inner workings driving the model's decision processes \cite{Gilpin2018ExplainingLearning}. Driven by the need for trustworthiness and reliability \cite{Lipton2016TheInterpretability}, numerous techniques, ranging from gradient-based \cite{Simonyan2013DeepMaps}, perturbation-based \cite{Ribeiro2016WhyClassifier} and contrastive approaches \cite{Abhishek2022Attribution-basedReview}, have emerged to assess \textit{a posteriori} (post-hoc) the behavior of opaque models \cite{Samek2021ExplainingApplications}. Within the branch of \textit{visual explanations}, \textit{saliency} methods aim to discriminate and identify relevant regions in the input space that highly excite the network and strongly influence the network predictions.

As successful state-of-the-art vision tasks' architectures commonly incorporate spatial convolution mechanism, \textit{Class Activation Maps} (CAM) \cite{Zhou2015LearningLocalization} have emerged as a popular and widely adopted technique for generating \textit{saliencies} 
that leverage the spatial information captured by convolutional layers. CAM(s) are computed by inspecting the feature maps and produce per-instance, class-specific attention heat maps that highlight important areas in the original image that drove the classifier. Building on this notion, Gradient-weighted CAM (Grad-CAM) \cite{Selvaraju2016Grad-CAM:Localization} and its variants, extend the original formulation by computing the linear weights from the averaged backpropagated gradients \wrt target class of each feature map. This generalization enables the use and application of the method without any modification or auxiliary training to the model.
Historically, naïve vanilla gradients have been cardinal in the development and evolution of \textit{saliency maps} \cite{Simonyan2013DeepMaps}; however input-gradients techniques (\eg output gradients \wrt inputs) quickly evolved to address the gradient saturation problem \cite{Shrikumar2017LearningDifferences, Rakitianskaia2015MeasuringNetworks, Sundararajan2017AxiomaticNetworks}, where the gradients of important features results in small magnitudes due to the model's function flattening in the vicinity of the input, misrepresenting the feature importance \cite{Sundararajan2016GradientsCounterfactuals}. Within the context of gradient visualizations, several \textit{counterfactual}-based works have been proposed in an attempt to address the saturation issue by feature scaling \cite{Sundararajan2017AxiomaticNetworks}, contribution decomposition \cite{Shrikumar2017LearningDifferences} and relevance propagation \cite{Bach2015OnPropagation}. In this direction, the insensitivity of \textit{baseline-methods} to their reference parameter \cite{Sundararajan2018AValues, Adebayo2018LocalValues} has spurred an area of research dedicated to baseline determination \cite{Ancona2017TowardsNetworks, Kindermans2017TheMethodsb,Yeh2019OnExplanations}. Since the introduction of the original proposition of CAM and Grad-CAM, several gradient-based techniques have been proposed in an effort to improve localization \cite{Shi2020Zoom-CAM:Labels, Jiang2021LayerCAM:Localization}, multi-instance detection \cite{Chattopadhay2017Grad-CAM++:Networks}, saliency resolution \cite{Qiu2023Fine-GrainedNetworks, Draelos2020UseNetworks}, noise and attribution sparsity \cite{Omeiza2019SmoothModels} and axiomatic attributions \cite{Fu2020Axiom-basedCNNs}. Despite numerous techniques being presented to address the \textit{saturation} phenomena, modern and widely adopted gradient-weighted CAM approaches still rely on vanilla gradients, which are inherently prone to gradient saturation. A recent work, namely Integrated Grad-CAM \cite{Sattarzadeh2021IntegratedScoring}, has been proposed aimed to address this issue which combines two well-established techniques: Integrated Gradients and Grad-CAM. Nonetheless, this method retains the same shortcomings as its underlying parent approach, that is, its lack of sensitivity to its baseline parameter, which underestimates contributions that align with its baseline.

\begin{figure}
  \centering
    % --- fake entry
    \begin{subfigure}[b]{0.1\textwidth}
        \captionlistentry{}
        \label{subfig:loc-0}
    \end{subfigure}
    \begin{subfigure}[b]{0.1\textwidth}
        \captionlistentry{}
        \label{subfig:loc-1}
    \end{subfigure}
    % --- fake entry

    \footnotesize
    \def\svgwidth{\columnwidth}
    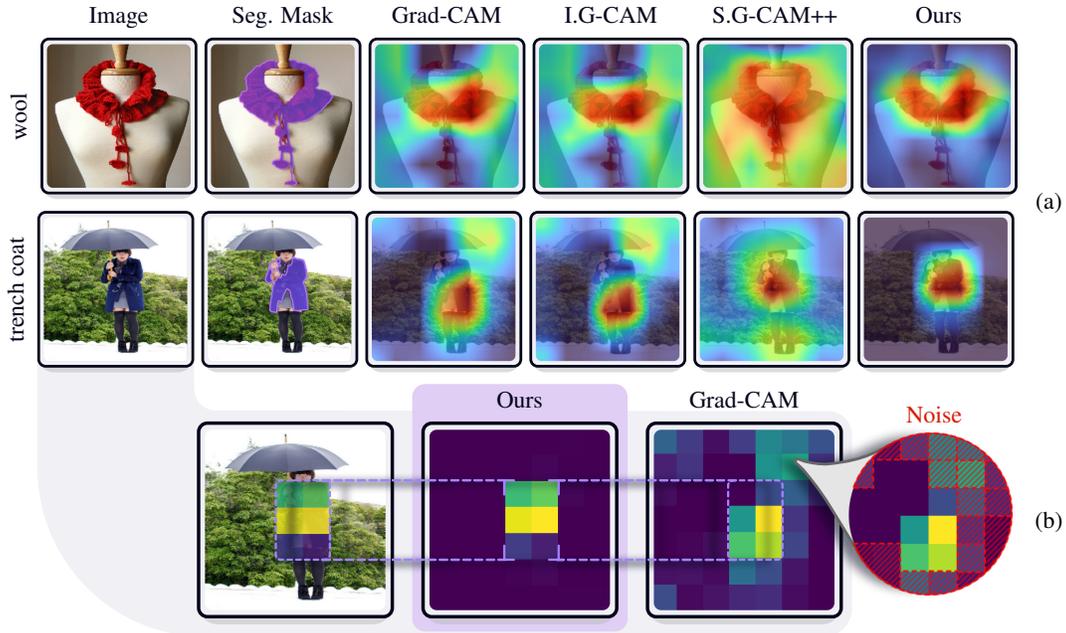

    \normalsize

   \caption{Explanatory functions on VGG-16 across samples from ImageNet-1k \cite{Russakovsky2014ImageNetChallenge}. Our approach produces sharper (less noisy) and higher localized heat maps with lower complexity than existing methods (\ref{subfig:loc-0}). \Cref{subfig:loc-1} shows the coarse heat map with respect to our method and baseline Grad-CAM \cite{Selvaraju2016Grad-CAM:Localization}.}
   \label{fig:frontback-hero-image}
\end{figure}

Following this research trajectory, we theorize and demonstrate that we can 
\textit{concurrently} improve four explanation quality \textit{key desiderata} in the context of human-interpretable saliencies \cite{Hedstrom2022Quantus:Beyond, Hedstrom2023TheMetaQuantus}: (i) fidelity, (ii) robustness, (iii) localization, and (iv) complexity. In this paper, we demonstrate that the explanations generated by our approach \textit{simultaneously} satisfy many desirable xAI properties by producing saliencies that are highly concentrated (\ie high \textit{localization}$\uparrow$) on the least number (low \textit{complexity}$\downarrow$) of stable (low sensitivity to infinitesimal perturbation) robust features (features which are consistently used). Our experiments reveal that \textit{Expected Grad-CAM} significantly outperforms currently state-of-the-art gradient- and non-gradient-based CAM methods across the tested xAI metrics in a large evaluation study. The results are consistent across different open image datasets. Qualitatively, our technique constructs \textit{saliencies} that are sharper (less noisy) and more focused on salient class discriminative image regions, as illustrated in \cref{fig:frontback-hero-image}. \Cref{fig:noise-cmp} shows that the saliency maps of popular gradient-based CAM methods are often noisy and appear sparse and uninformative \cite{Kim2019WhyMaps}, with large portions of pixels outside the relevant subject. In contrast, Expected Grad-CAM highlights only those features that systematically are utilized not only for a given sample but also for all the samples in its vicinity in the input space and thus, that produce the same prediction (\ie \textit{relative input stability}\cite{Agarwal2022RethinkingExplanations}).

\begin{figure}
  \centering
    % --- fake entry
    \begin{subfigure}[b]{0.1\textwidth}
        \captionlistentry{}
        \label{subfig:saturation-cmp-0}
    \end{subfigure}
    \begin{subfigure}[b]{0.1\textwidth}
        \captionlistentry{}
        \label{subfig:saturation-cmp-1}
    \end{subfigure}
    % --- fake entry

    \footnotesize
    \def\svgwidth{\columnwidth}
    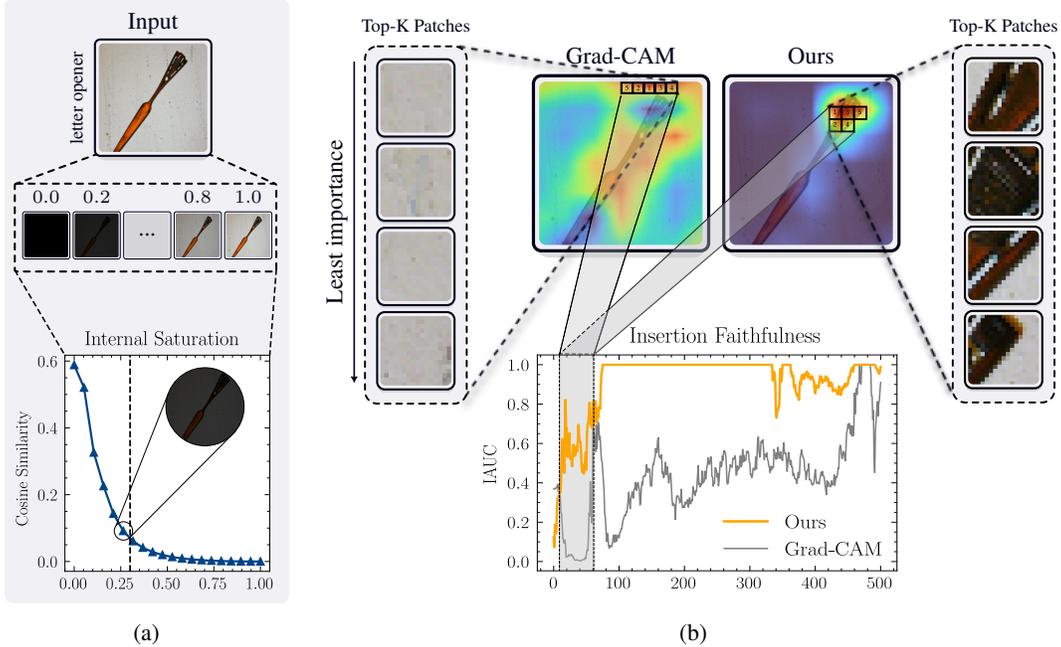

    \normalsize

    \caption{Comparison of attribution maps under internal saturation conditions. \Cref{subfig:saturation-cmp-0} illustrates the cosine similarity of the target layer's embeddings with respect to the interpolator parameter ($\alpha$) (see Appendix \ref{sec:internal-saturation} for more details). \Cref{subfig:saturation-cmp-1} displays the attribution maps of various methods under saturation conditions. Internal saturation causes the baseline method to under-represent feature importances across saturating ranges. By extracting the top-4 most important features (\Cref{subfig:saturation-cmp-1}), it is evident that the baseline method fails to capture relevant discriminative regions, resulting in low insertion AUCs (\Cref{subfig:saturation-cmp-1}) as these regions are not deemed important by the model.}

   \label{fig:saturation-cmp}
\end{figure}

\begin{figure}
  \centering
  
    \includegraphics[width=1\textwidth]{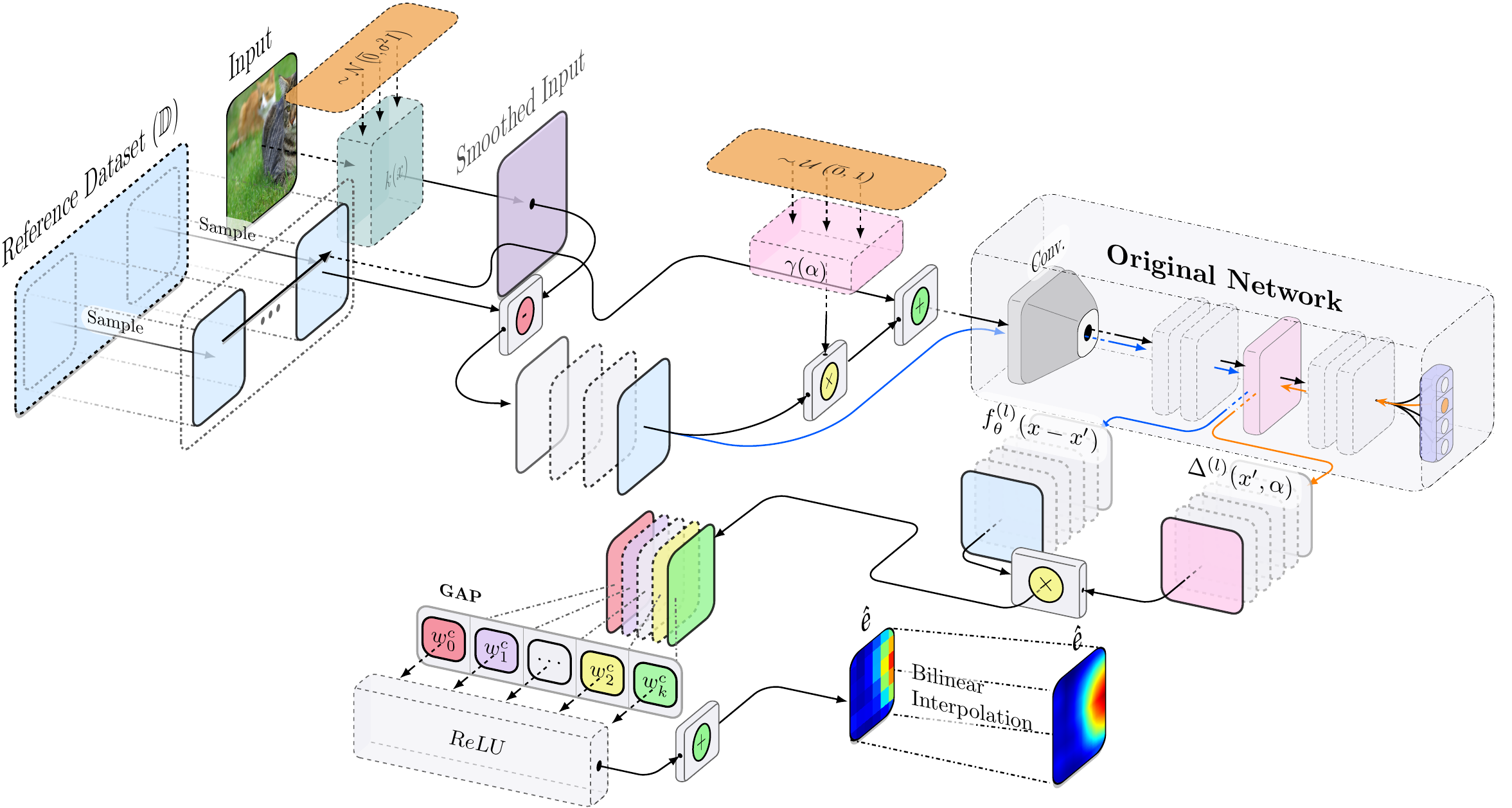}
  \caption{Overview of the proposed Expected Grad-CAM method. Given an input image, a target class, and a reference distribution to sample from, the class-discriminative explanation $\hat{e}$ is computed through input kernel smoothing and difference-from-reference comparisons.}

   \label{fig:method-diagram}
\end{figure}

Our method, \textit{Expected Grad-CAM}, tackles the current limitations of existing methods by reshaping the original gradient computation by incorporating the provably and well-established \textit{Expected Gradients} \cite{Erion2019ImprovingGradients} difference-from-reference approach followed by a smoothing kernel operator (\cref{fig:method-diagram}). As opposed to prior methods, our work solves the underestimation of the feature attribution (\cref{fig:saturation-cmp}) without introducing undesired side effects (\ie parameter insensitivity) by sampling the baseline parameter of the path integral from a reference distribution. As generated CAM(s) are coarse attention maps (\ie inherently low complexity saliencies), they are often used with the end goal of \textit{human-centered} interpretability of the function predictor and its behavior \cite{Alvarez-Melis2018OnMethods}. Therefore, it is crucial that such attribution methods highlight only stable features focusing only on salient areas of the original input \cite{Alvarez-Melis2018OnMethods}.

We summarize our contributions as follows: first, we provide a general scoring scheme that is not bound along a monotonic geometric path for generating gradient class activation maps that minimize infidelity for arbitrary perturbations. Second, we propose Expected Grad-CAM, a gradient-weighted CAM augmentation that produces class-specific heat maps that simultaneously improve four modern key explanatory quality desiderata. Third, we evaluate the effectiveness of our approach on a large evaluation study across $19$ quality estimators on recent explanation quality groupings \cite{Hedstrom2022Quantus:Beyond, Hedstrom2023TheMetaQuantus} \ie (i) Faithfulness, (ii) Robustness, (iii) Complexity, and (iv) Localization. Lastly, we demonstrate that our technique significantly outperforms state-of-the-art gradient- and non-gradient-based CAM methods.

%++++++++++++++++++++++++++++++++++++
% ++++++++++++++++++++++++++++++++++++

\section{Related work}
\label{sec:related-works}
%---------------------------------

The following section presents a brief discussion of prior attribution methods alongside their known shortcomings and notation.

\paragraph{Gradient-based explanations}
 This set of techniques encompasses the involvement of the neural network's gradients as a function approximator, translating complex nonlinear models into local linear explanations. These explanations are often encoded as attention heat maps, also known as saliencies. The cornerstone method within this category is \textbf{Input-Gradients} (vanilla gradients) \cite{Simonyan2013DeepMaps}.
Consider a classical supervised machine learning problem, where \( x \in \mathbb{R}^D \) be an input sample point for a neural network \( F: \mathbb{R}^D \rightarrow \mathbb{R}^C \). The class-specific input-gradients, are the backpropagated gradients of the output \wrt the input sample and are defined as:
\begin{equation}
    \phi_{i}(x;F^c) = \nabla_{x} F^c(x)
\end{equation}
where \( \phi_{i}(x;F^c) \) denotes the gradient of the class \( c \) \wrt the input \( x \). Notably, while not relevant to our approach, the feature visualization produced by deconvolution \cite{Zeiler2013VisualizingNetworks} and guided backpropagation \cite{Springenberg2014StrivingNet} are also tightly linked, with the latter letting flow only non-negative gradients.

\paragraph{Counterfactual explanations}
As gradients only express local changes, their utilization misrepresents feature importances across saturating ranges \cite{Sundararajan2017AxiomaticNetworks}. This class of methods tackles this issue by multiple nonlocal comparisons against a perturbed baseline by feature re-scaling \cite{Sundararajan2017AxiomaticNetworks}, blurring \cite{Fong2017InterpretablePerturbation}, activation differences \cite{Shrikumar2017LearningDifferences}, noise \cite{Smilkov2017SmoothGrad:Noise} or inpainting \cite{Alipour2022ExplainingSpaces}. Here, we primarily focus on two kinds of methods that are highly related to our work: Integrated Gradients \cite{Sundararajan2017AxiomaticNetworks} and SmoothGrad \cite{Smilkov2017SmoothGrad:Noise}.

\paragraph{Integrated Gradients} This method involves the summation of the (interior) gradients along the path of counterfactuals \cite{Sundararajan2017AxiomaticNetworks, Sundararajan2016GradientsCounterfactuals}. It is defined as:
\begin{equation}
    \phi_i^{IG}(x, x^\prime; F^c)=\left({x}-{x}^\prime\right) \int_{\alpha=0}^1 \nabla_{{x}} F^c\left({x}^\prime+\alpha\left({x}-{x}^\prime\right)\right) d \alpha
\end{equation}
where \( x \) is the input sample, \( x' \) is a given baseline, \( F^c \) is the neural network output for class \( c \), and \( \alpha \) is a scaling parameter that interpolates between the baseline and the input according to a given interpolation function $\gamma$ \cite{Sundararajan2017AxiomaticNetworks}.

\textbf{SmoothGrad:} This method addresses saliency noise caused by sharp fluctuations of gradients at small scales, due to rapid local variation in partial derivatives \cite{Smilkov2017SmoothGrad:Noise}, by denoising using a smoothing Gaussian kernel. It is defined as:
\begin{equation}
    \phi_i^{SG}(x; F^c)=\frac{1}{n} \sum_1^n \nabla_{{x}} F^c\left(x+\mathcal{N}\left(\overline{0}, \sigma^2\right); F^c\right)
\end{equation}
where \( x \) is the input sample, \( \mathcal{N}(\overline{0}, \sigma^2) \) is Gaussian noise with mean 0 and variance \( \sigma^2 \), \( F^c \) is the neural network output for class \( c \), and \( n \) is the number of noisy samples averaged.

\paragraph{Class activation mapping} 
This set of attention methods generates explanations by exploiting the spatial information captured by the convolutional layers. Class activation maps are generated by computing the rectified sum of all the feature map's activations times its weights. Formally, consider a network's target convolutional layer output having size $s$ and let \( f_{x, y}^k \in \mathbb{R} \) be the activation of unit $k$ in the target convolution layer at spatial location $(x, y)$. Then the class activation map \cite{Zhou2015LearningLocalization} for class $c$ at spatial location $(x, y)$ are computed as
\begin{equation}
    M_{x, y}^c=\sum_k w_k^c f_{x, y}^k
\end{equation}
where \( w \in \mathbb{R}^k \) are the weights of the fully connected layer and the result of performing a global average pooling for unit $k$ is $\sum_{x, y} f^k_{x, y}$. The ReLU has been omitted for readability. By generalizing CAM is possible to avoid the model's architectural changes by reinterpreting the Global Average Pooling weighting factors as the backpropagated gradients of any target concept \cite{Selvaraju2016Grad-CAM:Localization}. Then weights of unit $k$ are defined as
\begin{equation}
   w_k^c=\frac{1}{n} \sum_x \sum_y \frac{\partial y^c}{\partial A_{x, y}^k}
   \label{eq:gradcam-weights}
\end{equation}
where \( y^c \) is the score for class \( c \) before the softmax, \( A_{x,y}^k \) is the activation at location \( (x, y) \) of unit \( k \) in the target convolutional layer, and \( n \) is the total number of spatial locations in the feature map (\ie \(s \times s \)). 

% -----
Notably, despite the perturbation of the subregions is performed with distinct different techniques, \textit{DeepLift} \cite{Shrikumar2017LearningDifferences}, $input \times gradient$, and \textit{SmoothGrad} \cite{Smilkov2017SmoothGrad:Noise} they all work under a similar setup of \textit{Integrated Gradients} \cite{Sundararajan2017AxiomaticNetworks} as shown in previous work \cite{Ancona2017TowardsNetworks}. For instance, \textit{SmoothGrad} can be formulated as the \textit{path integral} where the interpolator function samples a single point from a Gaussian distribution \ie $\epsilon_\sigma \sim \mathcal{N}\left(\overline{0}, \sigma^2 I\right)$: 
\begin{equation}
    \phi_i^{S G}\left(F^c, x ; \mathcal{N}\right)=\frac{1}{n} \sum_{j=1}^n\left(x+\epsilon_\sigma^j\right) \frac{\partial F^c\left(x+\epsilon_\sigma^j\right)}{\partial x}
\end{equation}
\section{Method}
\label{sec:method}
%%%%%%%%%%%%%%%%%%%%%%%%%%%%%%%%%%%

We first provide some formal definitions. Consider a classification problem where we have a \textit{black-box} model function $f: \mathbb{X} \mapsto \mathbb{Y}$ that maps a set of inputs $x$ from the input space $\mathbb{X}$ to a corresponding set of predictions in the output space $\mathbb{Y}$ such that $\boldsymbol{x} \in \mathbb{X}$ and $\hat{y} \in \mathbb{Y}$ where $x \in \mathbb{R}^D$. The neural network $f_\theta: \mathbb{R}^D \rightarrow \mathbb{R}^C$ is parameterized by $\theta$ with an output space of dimension of $C$ classes where $\theta$ are the outcome of the training process \ie producing a mapping $f(\boldsymbol{x} ; \theta)=\hat{y}$.

We now define a \textit{saliency} $S$ as \textit{local feature attribution} $\Phi$ which deterministically map the input vector $x$ to an explanation $\hat{e}$ given some parameters 
$\lambda$:
\begin{equation}
    S_{f_\theta}: \mathbb{R}^D \times \mathbb{F} \times \mathbb{Y} \mapsto \mathbb{R}^B = \Phi(\boldsymbol{x}, f, \hat{y} ; \lambda)=\hat{e}
\end{equation}

The explanation $\hat{e}$ highlights the \textit{discriminative} regions within the input space $\mathbb{X}$ which highly drove the classifier towards $\hat{y}$. Given the mapping $f(\boldsymbol{x} ; \theta)-\hat{y}$, the saliency seeks to encode the influence of the learned behavior modeled by the parameters $\theta$ with $\theta_0 \sim p\left(\theta_0\right), f_\theta \sim p(\mathbb{D})$ where $\mathbb{D}$ represents the underlying training distribution. Notably, the saliency is computed with an arbitrary dimensionality output shape $\mathbb{R}^B$ which may differ from the input space $\mathbb{X}$. This holds true for \textit{CAM} techniques where the output is the products of the subsequent layers up to the target layer $l$, which often is in a lower dimensionality space due to the applied spatial convolutional blocks. This ultimately involves a transformation $T$ that maps the lower dimensional map back to the original input space $\mathbb{X}$ with $T: \mathbb{R}^B \rightarrow \mathbb{R}^D$.

%++++++++++++++++++++++++++++++++++++
% ++++++++++++++++++++++++++++++++++++

\subsection{Reshaping gradient computation}
% \label{sec:method}
%%%%%%%%%%%%%%%%%%%%%%%%%%%%%%%%%%%

The original formulation of Grad-CAM involves the usage of vanilla gradients, which, by expressing local changes, under-represent feature importances across saturating ranges \cite{Sundararajan2016GradientsCounterfactuals}. Previous works have addressed the saturation phenomena within CAM(s) by implementing popular perturbation techniques \cite{Sattarzadeh2021IntegratedScoring, Omeiza2019SmoothModels}, however, introducing undesirable side effects \ie \textit{baseline insensitivity}\cite{Sundararajan2018AValues} or poor robustness and stability \cite{Alvarez-Melis2018OnMethods,Sundararajan2017AxiomaticNetworks, Ghorbani2017InterpretationFragile}. Expected Grad-CAM tackles the gradient saturation without interposing insensitivity to its baseline parameter while causally improving four key desiderata in the context of human-interpretable saliencies: (i) fidelity, (ii) robustness, (iii) localization and (iv) complexity. 
The augmentation operates only on the gradient computation and works under a similar setup as the provably Expected Gradients \cite{Erion2019ImprovingGradients} technique. Let a subset $S_k$ of $k$ features to be perturbed, then an universal, unconstrained scheme that constructs \cite{Ancona2017TowardsNetworks} interpolated inputs by replacement \cite{Yeh2019OnExplanations} can be defined as: 
\begin{equation}
    \mathbf{x}\left[\mathbf{x}_S=x^\prime\right]_j=
    x_j\mathbb{I}(j \notin S) + x^\prime \mathbb{I}(j \in S)
\end{equation}
where $\mathbb{I}$ is the indicator function and $x^\prime$ is a reference baseline. The gradient-scheme which iteratively identifies salient regions in the input space $\mathbb{X}$, can be extended as the path integral \cite{Sundararajan2017AxiomaticNetworks} which is sampled from a reference distribution (Expected Gradients) \cite{Erion2019ImprovingGradients}
\begin{equation}
   \Phi_i(f, x ; \mathbb{D})=\int_{x^{\prime}}\left(\phi_i^{I G}\left(f, x, x^{\prime}\right) p_D\left(x^{\prime}\right) d x^{\prime}\right) 
\end{equation}

More generally, we can present an augmentation with a similar scheme which is not bound along a monotonic geometric path, aimed at addressing baseline insensitivity 
\begin{definition}
Given a black-box model function $f_\theta$ and a generic meaningful perturbation $\boldsymbol{I}$ with $\boldsymbol{I} \sim p_D(\boldsymbol{I})$, where $f^{(l)}_\theta$ is the latent representation of $f_\theta$ at arbitrary layer $l$. Then any gradient-based CAM scoring mechanism $\Phi$ can be formulated as:
\begin{equation}
    \Phi(f, x)=\int_{\boldsymbol{I}} f_\theta^{(l)}(\boldsymbol{I I}^T) \, \phi^{IG}(f, \mathbf{x}, \boldsymbol{I})\, p_D(\boldsymbol{I}) \, d \boldsymbol{I}
\end{equation}
where $\phi^{IG}$ is any Integrated Gradients attribution framework,
\begin{equation}
    \phi^{IG}(f, x, \boldsymbol{I})=\int_{\alpha=0}^1 \nabla f(x+\boldsymbol{I}(\alpha-1)\, )\; d \alpha
\end{equation}
\end{definition}
\begin{remark}
As pointed out by previous work \cite{Yeh2019OnExplanations} $\Phi(f, x, \boldsymbol{I})$ can be replaced by any functional kernel which holds the \textit{completeness axiom} \cite{Sundararajan2017AxiomaticNetworks} \ie $f_\theta^{(l)}(\boldsymbol{I I}^T) \, \phi^{IG}(f, x, \boldsymbol{I})=f(x)-f(x-\boldsymbol{I})$ 
\end{remark}

\begin{remark}
If the construction of the perturbation matrix $\boldsymbol{I}$ occurs over a linear space and using a feature scaling policy over a single constant baseline, then the formula becomes equivalent to Integrated Grad-CAM \cite{Sattarzadeh2021IntegratedScoring}.
\end{remark}

Extending this notion, we can formulate a gradient-based local attribution scheme aimed at concurrently improving faithfulness and human-interpretable desirable properties by minimizing infidelity \cite{Yeh2019OnExplanations}
\begin{definition}
Given the black-box function model $f_\theta^{(l)}$ with $(l)$ being any intermediary layer and $\boldsymbol{\eta}$ any smoothing, distribution-preserving perturbation and $\boldsymbol{I}$ any meaningful perturbation with $\boldsymbol{\eta} \sim \mu_{\boldsymbol{\eta}}$ and $\boldsymbol{I} \sim \mu_{\boldsymbol{I}}$. Then the CAM gradient augmentation can be formulated as:
\begin{equation}
     \Phi(f, x, \boldsymbol{I})=
    \left(\int_{\boldsymbol{\eta}} f_\theta^{(l)}(\boldsymbol{\eta \,\eta}^T) d \mu_{\boldsymbol{\eta}}\right)^{-1}
    \int_{\boldsymbol{I}} f_\theta^{(l)}(\boldsymbol{I I}^T) \Phi(f, \mathbf{x}, \boldsymbol{I}) d \mu_{\boldsymbol{I}}
    % \nonumber
\end{equation}
\end{definition}

In this context, the original weighting factors (\cref{eq:gradcam-weights}) can be reformulated as the 
 unaltered linear combination of the smoothed, distribution-sampled, and perturbed integrated gradients\footnote{Note that varying the smoothing distribution alters the type of smoothing kernel applied; when $\eta \sim \mathcal{N}\left(\overline{0}, \sigma^2 \boldsymbol{I}\right)$ then the functional becomes SmoothGrad\cite{Smilkov2017SmoothGrad:Noise} }
\begin{equation}
    w_k^c=\frac{1}{Z} \sum_i \sum_j \Phi(f^{(l)}_C, x, \mathbf{I}, \boldsymbol{\eta})
\end{equation}

\subsection{Robust perturbations by data distillation}

Recent works have shown that \textit{faithfulness} and \textit{fidelity} are only one of many desirable properties a quality explainer should possess \cite{Hedstrom2022Quantus:Beyond}. In this direction, the choice of the perturbation is a key component to \textit{(i) preserve the sensitivity axiom} \cite{Sundararajan2017AxiomaticNetworks}, \textit{(ii) guarantee stability not solely at the input and output, but also \wrt to intermediary latent representations} \cite{Agarwal2022RethinkingExplanations} and \textit{(iii) ensure robustness to infinitesimal perturbations} \cite{Slack2019FoolingMethods, Moosavi-DezfooliUniversalPerturbations}. The usage of \textit{constant baselines} provides a weak notion of completeness which does not account for noise within the data \cite{Yeh2019OnExplanations}, ultimately showing \textit{high sensitivity} and \textit{high reactivity} to noise. Thence, we construct baselines that approximate a given reference distribution rather than a fixed static value with a similar integration scheme as the provably Expected Gradients technique \cite{Erion2019ImprovingGradients}. This allows to maintain a similar gradient \textit{smoothing behavior} introduced by the usage of a Gaussian kernel \cite{Smilkov2017SmoothGrad:Noise, Omeiza2019SmoothModels} but higher robustness as each intermediary sample is distilled with a perturbation $\mathbf{I}$ that is close to the underlying training data distribution. Ultimately allowing fewer intermediary samples to fall outside of the data distribution (OOD).

\begin{definition}
\label{prop:b}
Let $\Phi(f, x, \boldsymbol{I})$ be a local feature attribution scoring method with $\boldsymbol{I}$ a crafted perturbation, and the integral over the outer products of all possible perturbation is inversible $\left(\int \boldsymbol{I I}^T d \mu_{\boldsymbol{I}}\right)^{-1}$. Then the dot product between $\Phi$ and $\boldsymbol{I}$ must satisfy the completeness axioms

\begin{equation}
    \boldsymbol{I}^T \Phi(f, x, \boldsymbol{I}) = f^c(x) - f^c(x-\boldsymbol{I})
\end{equation}
\end{definition}

% to save space ---
Building on the above definition, we define a robust perturbation derived from the distillation of the underlying data distribution $\mathbb{D}$ using Monte Carlo sampling. The robust perturbation is given by the expectation $\mathbb{E}_{\boldsymbol{I}^{\prime} \sim \mathbb{D}, \alpha \sim U(0,1)} [\boldsymbol{I}]$, where $\boldsymbol{I} = x - \boldsymbol{I}^\prime$.

% ++++++++++++++++++++++++++++++++++++
% ++++++++++++++++++++++++++++++++++++

\subsection{Expected Grad-CAM and connection to path attribution methods}

By thus careful crafting of the perturbation matrix $\boldsymbol{I}$ using a smoothing, distribution-preserving kernel $k(x, \alpha)$ we can formulate Expected Grad-CAM. 

\begin{definition}
\label{def:egcam-weights}
Let $\boldsymbol{I}$ be a robust data distilling perturbation and $\boldsymbol{\eta}$ the results of a smoothing kernel with $\mu_{\boldsymbol{I}} \approx \mathbb{D}$ and $\boldsymbol{\eta} \sim \mu_{\boldsymbol{\eta}}$. Given that $\phi$ is any Integrated Gradients perturbation scheme, we define the \textbf{Expected Grad-CAM} weights  of unit $k$ as:
\begin{equation}
     w_k^c = \frac{1}{n} \sum_{i, j} \underset{\boldsymbol{I}^{\prime} \sim \mu_{\boldsymbol{I}}}{\mathbb{E}} 
     \left[
     % ---
     k(x)
    \int_{\boldsymbol{I}} f_\theta^{(l)}(\boldsymbol{I I}^T) \phi(f, \boldsymbol{x}, \boldsymbol{I}) d \mu_{\boldsymbol{I}}
     % ---
     \right]
\end{equation}
where,
\begin{equation}
k(x) = \left(\int_{\boldsymbol{\eta}} f_\theta^{(l)}(\boldsymbol{\eta \,\eta}^T)\, d \mu_{\boldsymbol{\eta}}\right)^{-1}
\end{equation}
Thus $k(x)$ is the first moment of all the smoothing functionals under the distribution $\mu_{\boldsymbol{\eta}}$ encoded at an arbitrary intermediary layer $(l)$ and $(i, j)$ the feature map spatial locations. Explicitly, $f_\theta^{(l)}$ produces layer-specific latent representations modeled under the learned behavior $\theta$. The smoothing kernel $k(x)$ directly controls the sensitivity of the explanation around the sample $x$, implicitly driving the complexity of the explanation. Whereas any smoothing kernel can be adopted, we found that $\mu_\eta \sim \mathcal{N}\left(\overline{0}, \sigma^2 \boldsymbol{I}\right)$ and $\mu_\eta \sim \mathcal{U}(0, 1)$ produces similar smoothing performances, therefore the latter has been used in the experiments.
\end{definition}

% =here ---

%++++++++++++++++++++++++++++++++++++
\begin{figure}
  \centering
    % --- fake entry
    \begin{subfigure}[b]{0.1\textwidth}
        \captionlistentry{}
        \label{subfig:noise-cmp-0}
    \end{subfigure}
    \begin{subfigure}[b]{0.1\textwidth}
        \captionlistentry{}
        \label{subfig:noise-cmp-1}
    \end{subfigure}
    % --- fake entry

    \scriptsize
    \def\svgwidth{\columnwidth}
    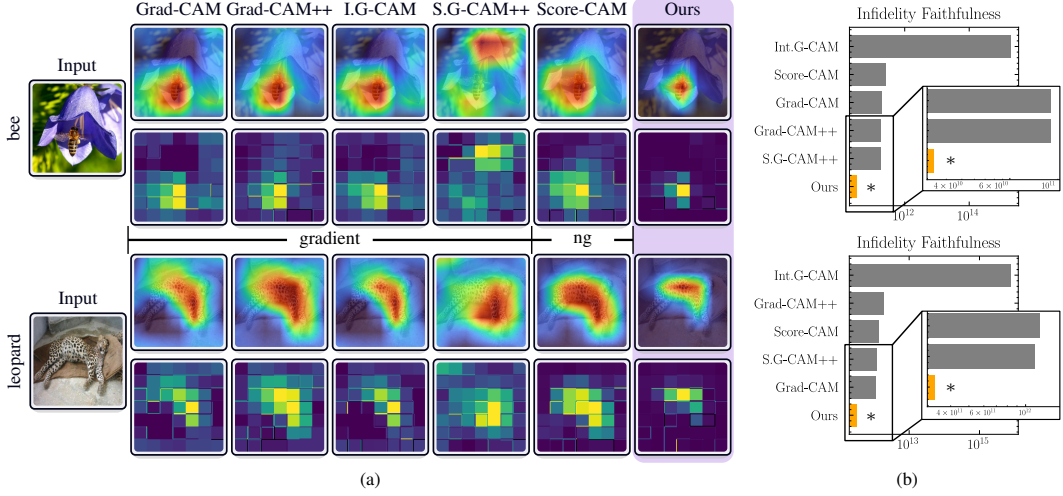

    \normalsize

    \caption{Comparison of saliencies generated by different gradient- and non-gradient-based methods. \Cref{subfig:noise-cmp-0} shows the superimposed (top row) and raw coarse saliencies (bottom row) generated by each method. \Cref{subfig:noise-cmp-1} presents the Infidelity scores \cite{Yeh2019OnExplanations} (using log-scale) for the different methods. While baseline methods are noisy with low localization, our method produces sharper, more localized explanations, outperforming even non-gradient-based techniques, and resulting in significantly lower infidelity scores (\cref{subfig:noise-cmp-1}).}

   \label{fig:noise-cmp}
\end{figure}
% ++++++++++++++++++++++++++++++++++++

\begin{definition}
Let $\xi$ be any matrix perturbation with $\xi \sim \mu_\xi$ and $\xi \xi^T$ produce a covariance matrix such that $\xi \xi^T = K_{\xi\xi}$. Then $\int \xi p(\xi) d\xi$ corresponds to the expectation of the perturbation $\xi$ under that distribution.
\begin{equation}
\mathbb{E}[\xi]=\int \xi \mu_{\xi}(\xi) \; d \xi
\end{equation}
\end{definition}  

\begin{definition}
\label{def:egcam-saliency}
Let $S_{f_\theta}^C$ the coarse saliency generated for the model $f_\theta$ upon class $C$ and $A^{k, (l)}$ the activations $k$ of an arbitrary layer $l$. Then the explanation $S_{f_\theta}^C$ is constructed as the unaltered linear combination of the smoothed perturbed expected gradients as follows:
\begin{equation}
    S_{f_\theta}^C = \operatorname{ReLU}\left(\sum_k^N w_k^c A^{k, (l)}\right)
\end{equation}
\end{definition}

As discussed by previous work \cite{Ancona2017TowardsNetworks} and introduced at the beginning of this paper, despite the perturbations are crafted with different distinct techniques and thus are more consequential and insightful to discuss unconstrained non-monotonic perturbations, such schemes can always be re-formulated under a geometrical path integral method setup. That is, given the universal formula for a path method \cite{Sundararajan2017AxiomaticNetworks} $\phi_i^\gamma(f, x)$ and its interpolator function $\gamma$ as:

\begin{equation}
    \phi_i^\gamma(f, x)=\int_{\alpha=0}^1 \frac{\partial F(\gamma(\alpha))}{\partial \gamma_i(\alpha)} \frac{\partial \gamma_i(\alpha)}{\partial \alpha} d \alpha \quad\quad \text{ where } \qquad
    \begin{aligned}
    &\gamma(0):=x^{\prime},\\
    &\gamma(1):=x
    \end{aligned}
\end{equation}

Given a linear interpolation path such that the one employed in IG \cite{Sundararajan2017AxiomaticNetworks}:
\begin{equation}
    \gamma^{I G}(\alpha)=x^{\prime}+\alpha \times\left(x-x^{\prime}\right) \qquad\qquad\qquad \text { where } \qquad \alpha \in[0,1]
\end{equation}

\begin{definition}
\label{def:ig-path-integral}
Let $y^c_{\gamma(\alpha)}$ the class-specific model's output at the interpolated point $\alpha$ given the interpolator function $\gamma$. Then the linear Path Integral formulation of the \textbf{Expected Grad-CAM} weights at spatial location $(i, j)$ can be denoted as:
\begin{equation}
    \begin{aligned}
    w_k^c&=\underset{x^{\prime} \sim \mathbf{D}, \alpha \sim U(0,1)}{\mathbb{E}}\left[k(x, \alpha) \frac{\partial y_{\gamma(\alpha)}^c}{\partial A_{i, j}^{k, (l)}} \frac{\partial \gamma(\alpha)}{\partial \alpha}\right] \\
    &=\underset{x^{\prime} \sim \mathrm{D}, \alpha \sim U(0,1)}{\mathbb{E}}k(x, \alpha)\left[f_\theta^{(l)} \left(x_{i, j}-x_{i, j}^{\prime}\right) \Delta_{i,j}^{k, (l)}(x^\prime, \alpha)\right] \\
   &=\frac{1}{n} \sum_{s=1}^n k(x, \alpha) \left[f_\theta^{(l)} \left(x_{i, j}-x_{i, j}^{\prime, s}\right) \Delta_{i, j}^{k, (l)}(x^{\prime, s}, \alpha^s) \right]  
    \end{aligned}
\end{equation}
where,
\begin{equation}
    \Delta_{i, j}^{k, (l)}(x^\prime, \alpha) = \frac{\partial f^c_\theta\left(x^{\prime}+\alpha\left(x-x^{\prime}\right)\right)}{\partial A_{i, j}^{k, (l)}}
\end{equation}
represents the partial derivative of the class $c$ output with respect to the activation $A_{i, j}^{k, (l)}$, evaluated at the interpolated point $x^{\prime} + \alpha(x - x^{\prime})$ for a baseline $x^\prime$ sampled from a distribution $\mathrm{D}$.

\end{definition}

\section{Experiments}
\label{sec:experiments}
%%%%%%%%%%%%%%%%%%%%%%%%%%%%%%%%%%%%%%%%%%%%%%%%%%

%-------------------------
\begin{figure}
    \centering
    % --- fake entry
    \begin{subfigure}[b]{0.1\textwidth}
        \captionlistentry{}
        \label{subfig:qual-eval-0}
    \end{subfigure}
    \begin{subfigure}[b]{0.1\textwidth}
        \captionlistentry{}
        \label{subfig:qual-eval-1}
    \end{subfigure}
    % --- fake entry
    
  \scriptsize
    \def\svgwidth{\columnwidth}
    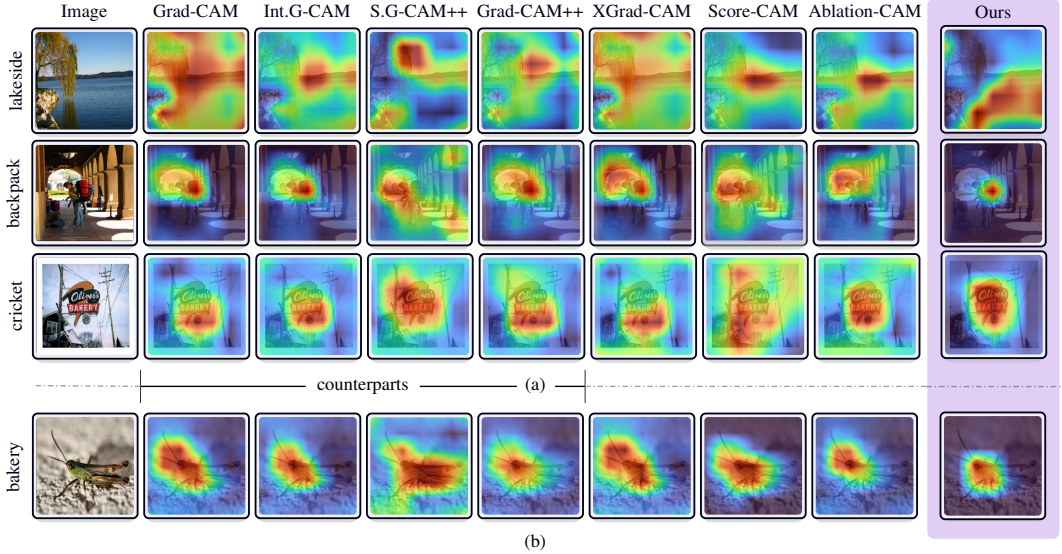

  \normalsize
  
   \caption{Comparison of attribution maps for various methods under normal conditions (\ref{subfig:qual-eval-1}) and internal saturation conditions (\ref{subfig:qual-eval-0}). Our method (\textit{Expected Grad-CAM}) provides sharper, more localized, and more stable explanations compared to its direct counterparts, namely \textit{Grad-CAM} \cite{Selvaraju2016Grad-CAM:Localization}, \textit{Grad-CAM++} \cite{Chattopadhay2017Grad-CAM++:Networks}, \textit{Smooth Grad-CAM++} \cite{Omeiza2019SmoothModels}, and \textit{Integrated Grad-CAM} \cite{Sattarzadeh2021IntegratedScoring}. Additionally, it offers competitive explanations compared to non-gradient and more complex methods through gradient augmentation. For a complete comparison, refer to \Cref{sec:supp-qualitative}.}

  \label{fig:qualitative-evaluation}
\end{figure}
% %-------------------------

In line with previous works \cite{Jiang2021LayerCAM:Localization, Wang2019Score-CAM:Networks, Jiang2021LayerCAM:Localization} we evaluate our proposed method quantitatively and qualitatively.

\begin{bsect}{Datasets}
We consider the \textit{ILSVRC2012} \cite{Russakovsky2014ImageNetChallenge}, \textit{CIFAR10} \cite{Ho-Phuoc2018CIFAR10Humans} and \textit{COCO} \cite{Lin2014MicrosoftContext} with images of size $224 \times 224$. The first two datasets have been used across the quantitative metrics, while the latter only for the localization evaluations, where the segmentation masks of each sample have been employed.
\end{bsect}\\
\begin{bsect}{Models}
Each metric is evaluated across popular feed-forward CNN architecture. In style with prior literature, we restricted our analysis to \textit{VGG16} \cite{Simonyan2014VeryRecognition}, \textit{ResNet50} \cite{He2015DeepRecognition} and \textit{AlexNet} \cite{Krizhevsky2012ImageNetNetworks}. In all cases, the default pre-trained \textit{PyTorch} torchvision implementation has been adopted.
\end{bsect}\\
\begin{bsect}{Metrics}
In contrast to prior works, we comprehensively evaluate our technique across an extensive set of traditional and modern metrics. We provide a full characterization of the behavior of our method by evaluating not just faithfulness, but rather all the different explanatory qualities across recent explanation quality grouping \cite{Hedstrom2022Quantus:Beyond} \ie (i) Faithfulness, (ii) Robustness, (iii) Complexity, and (iv) Localization. In \Cref{table:nomenclature} are presented all the evaluated metrics categorized by quality groupings, while the extended quantitative results are available in \Cref{sec:supp-quantitative}.
\end{bsect}\\
\begin{bsect}{Baselines}
We compare our proposed technique against recent and relevant methods including Grad-CAM \cite{Selvaraju2016Grad-CAM:Localization}, Grad-CAM++ \cite{Chattopadhay2017Grad-CAM++:Networks}, Smooth Grad-CAM++ \cite{Omeiza2019SmoothModels}, Integrated Grad-CAM \cite{Sattarzadeh2021IntegratedScoring}, HiRes-CAM \cite{Draelos2020UseNetworks}, XGrad-CAM \cite{Fu2020Axiom-basedCNNs}, LayerCAM \cite{Jiang2021LayerCAM:Localization}, Score-CAM \cite{Wang2019Score-CAM:Networks} and Ablation-CAM \cite{Desai2020Ablation-CAM:Localization}
\end{bsect}

\paragraph{Qualitative evaluations}
\label{sec:qualitative-evaluation}

\begin{figure}
  \centering
    % --- fake entry
    \begin{subfigure}[b]{0.1\textwidth}
        \captionlistentry{}
        \label{subfig:debug-cmp-0}
    \end{subfigure}
    \begin{subfigure}[b]{0.1\textwidth}
        \captionlistentry{}
        \label{subfig:debug-cmp-1}
    \end{subfigure}
    % --- fake entry

    \scriptsize
    \def\svgwidth{\columnwidth}
    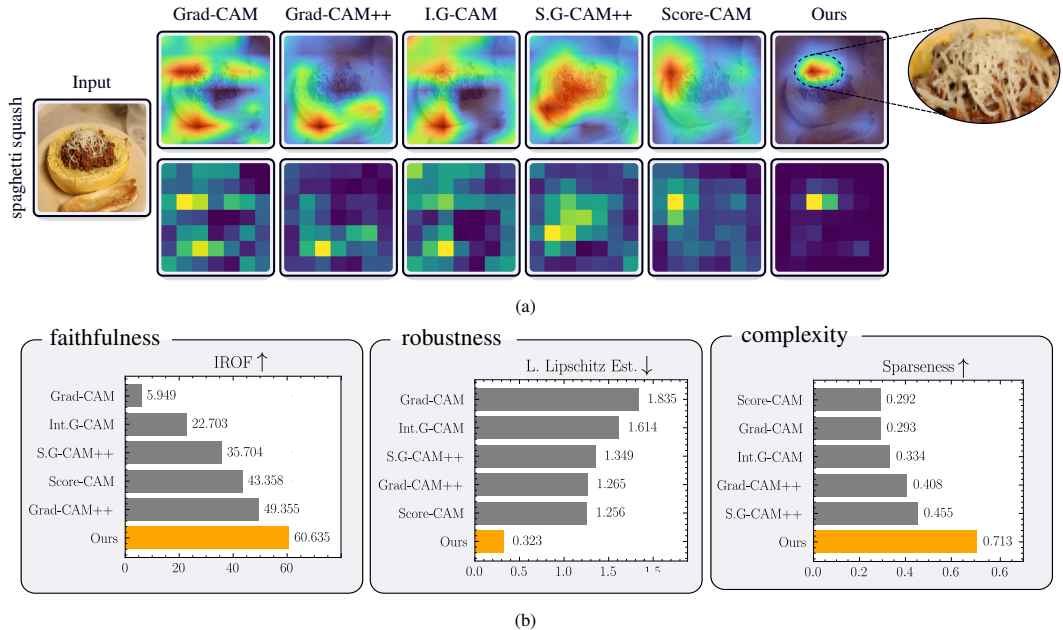

    \normalsize
    
    \caption{Comparison of saliencies generated by different gradient- and non-gradient-based methods. \Cref{subfig:debug-cmp-0} shows the superimposed (top row) and raw coarse saliencies (bottom row) generated by each method. Our method consistently produces more focused and sharper saliencies compared to both gradient-based and non-gradient-based methods (e.g., Score-CAM). \Cref{subfig:debug-cmp-1} demonstrates that our approach concurrently improves key xAI properties: (i) faithfulness, (ii) robustness, and (iii) complexity, significantly outperforming even non-gradient-based methods.}
    
   \label{fig:debug-cmp}
\end{figure}

In \Cref{fig:qualitative-evaluation} we present an excerpt of the explanations generated during the computation of the quantitative evaluations on the \textit{ILSVRC2012} validation set. By inspecting the attribution sparsity and localization characteristics of each explanation, our method (\textit{Expected Grad-CAM}), generally produces saliencies that are more localized and focused on the attuned human-centric understanding of the composition of the attributes of the labels. An explanation designed for human fruition \ie aimed at building the model's trustworthiness should be encoded as such to not \textit{disrupt trust}; this implies that an \textit{human-interpretable} explanation should be restricted to the most important pertinent and stable features: it should contains the least number of stable features which do maximally fulfill the notion of fidelity (\cref{fig:debug-cmp,fig:noise-cmp}). In \Cref{subfig:qual-eval-0} it is observed qualitatively that every other compared attribution method breaks such condition: given the labels \textit{lakeside} and \textit{backpack} the explanations highlights areas which are not pertinent with label-related attributes \ie the sky and portions of the tree (\cref{subfig:qual-eval-0}) and parts of the background (\cref{subfig:qual-eval-1}) respectively.

%-------------------------
% \input{tables/agg_table}
\begin{table}
\caption{\textit{Faithfulness}, \textit{Robustness} and \textit{Complexity Metrics}. Values evaluated on ILSVRC2012\cite{Russakovsky2014ImageNetChallenge} on VGG16 \cite{Simonyan2014VeryRecognition}. Extended results are available in \Cref{sec:supp-quantitative}.}
\label{table:agg-table}

\footnotesize
   \centering
   \begin{tabular}{@{}llcccccccc@{}}
    \\
        \toprule
        &&\multicolumn{3}{c}{Faithfulness} &\multicolumn{3}{c}{Robustness} &\multicolumn{2}{c}{Complexity}\\
        \cmidrule(r){3-5}
        \cmidrule(r){6-8}
        \cmidrule(r){9-10}
        &\multicolumn{1}{l}{\textbf{Method}} & \multicolumn{1}{c}{\textbf{$\downarrow$ P.F.}}
        & \multicolumn{1}{c}{\textbf{$\uparrow$ Suff.}}
        & \multicolumn{1}{c}{\textbf{$\downarrow$ Inf.}}
        & \multicolumn{1}{c}{\textbf{$\downarrow$ L. Est.}} & \multicolumn{1}{c}{\textbf{$\downarrow$ M. Sens.}}
        
        &\multicolumn{1}{c}{\textbf{$\downarrow$ A. Sens.}} & \multicolumn{1}{c}{\textbf{$\downarrow$ CP.}} & \multicolumn{1}{c}{\textbf{$\uparrow$ SP.}}\\
        \midrule
        
        &Grad-CAM  & 55.36 & 1.91 & 8.12 & 0.38 & 0.27 & 0.20 & 10.56 & 0.38  \\
        &Grad-CAM++ & 56.93 & 1.87 & 7.98 & 0.32 & \textbf{0.192} & 0.15 & \underline{10.53} & 0.40 \\
        &Sm. Grad-CAM++ & 56.38 & 1.89 & 7.50 & 0.51 & 0.51 & 0.27 & 10.60 & 0.35\\
        &Int. Grad-CAM & 57.36 & 1.83 & 8.92 & 1.05 & 1.00 & 1.00 & 10.59 & 0.36\\
        \midrule
        &HiRes-CAM & 57.49 & 1.74 & \underline{5.73} & 0.99 & 1.00 &1.00 & 10.54 & \underline{0.40}\\
        &XGrad-CAM & 57.32 & \underline{1.98} & 7.88 & 0.37 & 0.23 & 0.18 & 10.56 & 0.38 \\
        &LayerCAM & \underline{58.15} & 1.74 & 7.22 & \underline{0.31} & 0.19 & \textbf{0.14} & 10.56 & 0.38\\
        &Score-CAM & 5.37 & 1.91 & 7.39 & 0.68 & 0.65 &0.53 & 10.56& 0.38\\
        &Ablation-CAM & 57.36 & 1.83 & 7.28 & 1.05 & 1.00 & 1.00 & 10.59 & 0.36\\
        \midrule 
           
        &Expected Grad-CAM & \textbf{62.39} & \textbf{2.10} & \textbf{4.99} & \textbf{0.24} & \underline{0.194} & \underline{0.15} & \textbf{10.43} & \textbf{0.47}\\
        % ---
       \bottomrule
   \end{tabular}
   
\normalsize
\end{table}
%-------------------------

\paragraph{Quantitative evaluations}
\label{sec:quantitative-evaluation}

Following, we assess the validity of our claims quantitatively across various desirable explanatory qualities. The extended quantitative results are available in \Cref{sec:supp-quantitative}.

\begin{bsect}{Faithfulness}
Examining traditional \textit{faithfulness} metrics (Insertion and Deletion AUCs) across popular benchmarking networks on a large chunk of \textit{ILSVRC2012}, showed promising results (\cref{table:ins-del-ext}). Our method \textit{Expected Grad-CAM}, outperformed jointly its gradient and non-gradient-based counterparts as well as more advanced variation of CAM, which do not solely rely on a gradient augmentation, in both the \textit{insertion} and \textit{deletion} aspects. Towards a more comprehensive comparison, we then verified our technique against more recent metrics. Unsurprisingly, \textit{IROF} \cite{Rieger2020IROF:Methods} and \textit{Pixel Flipping \cite{Bach2015OnPropagation}} (\cref{table:agg-table}) showed agreement with traditional metrics as they fundamentally assess similar explanatory qualities. Our technique scored higher than others on the \textit{Sufficiency} \cite{Dasgupta2022FrameworkExplanations} metric, due to greater \textit{stability} and \textit{robustness} (\cref{table:agg-table}). Finally, we tested \textit{Expected Grad-CAM}'s performances in terms of \textit{infidelity} \cite{Yeh2019OnExplanations}, which, expectedly showed the highest results. For fairness, we provided also results with respect to known metrics that produce disagreeing rank-order results \cite{Rong2022AMethods, Hedstrom2022Quantus:Beyond, Hedstrom2023TheMetaQuantus} \ie \textit{Faithfulness Estimate} \cite{Alvarez-Melis2018TowardsNetworks} where our approach was the second best scoring explainer. 
\end{bsect}\\
\begin{bsect}{Stability}
In \cref{table:robustness-ris-ros} are presented the results \wrt to the relative- input and out stability metrics. our method showed the lowest score overall (\textit{highest stability}), while achieving best or second-best robustness scores (\cref{table:agg-table}).
\end{bsect}

\section{Conclusion and broader impact}
\label{sec:conclusion}
% -------------------------

In this paper, we advanced current CAM's gradient faithfulness by proposing \textit{Expected Grad-CAM} which simultaneously addresses the saturation and sensitivity phenomena, without introducing undesirable side effects. Revisiting the original formulation as the smoothed expectation of the perturbed integrated gradients, one can concurrently construct more faithful, localized, and robust explanations that minimize infidelity. 
Despite qualitative assessment being highly subjective, quantitative evaluations are also teeming with indeterminate, ambiguous results that span further than the rank-order issues. While faithfulness is a universally desirable underlying explanatory quality, individual metrics, which do assess such property, only define a distinct notion of such a multifaceted trait, potentially delineating unwanted aspects. While careful modulation of the smoothing functional allows for fine-grained control of the complexity characteristic of the explanation, where, through sensitivity reduction, produces more human-interpretable saliencies; it contrastingly influences the current notions of faithfulness. Perhaps, further adaption of existing metrics may be necessary to embody human-interpretability; nevertheless, existing qualitative and quantitative assessments proved the superiority of our approach.

\label{sec:broader-impact}
\begin{bsect}{Broader impact}
This paper highlights the value and effectiveness of Expected Grad-CAM in comparison to current state-of-the-art approaches across a comprehensive set of modern evaluation metrics. We demonstrated that our technique satisfies many desirable xAI properties by producing explanations that are highly concentrated on the least number of stable robust, features. Our experiments revealed that many state-of-the-art approaches underperform on modern metrics. 
Ultimately, as our technique is intended to replace the original formulation of Grad-CAM, we hope new and existing approaches will build on it.
\end{bsect}

% +++++++++++++++++++++++++++++++++++++++++++++++++++++
% +++++++++++++++++++++++++++++++++++++++++++++++++++++

% sections ----------------------------------

\bibliographystyle{plainnat}
\bibliography{references}

%%%%%%%%%%%%%%%%%%%%%%%%%%%%%%%%%%%%%%%%%%%%%%%%%%%%%%%%%%%%

\newpage
\appendix

\section{Appendix}
\label{sec:supplementary-material}
% ------------------------------------

Following are presented the extended results and remarks about notation and nomenclature. In \Cref{table:nomenclature} are listed the evaluated abbreviated metric names followed by their source, categorized by the underlying explanatory quality they seek to assess \cite{Hedstrom2022Quantus:Beyond}. Where applicable, IG-CAM and SG-CAM abbreviation have been used in place of Integrated Grad-CAM \cite{Sattarzadeh2021IntegratedScoring} and Smooth Grad-CAM++ \cite{Omeiza2019SmoothModels} respectively. All results have been computed on a single A100-SXM4 80GB platform and a Xeon Gold 5317 with CUDA v12.0.

% nomenclature --------------
\begin{table}[htb]
   \caption{Nomenclature of all the evaluated metrics and their source }
   \label{table:nomenclature}
    \newcommand{\ld}[1]{\footnotesize#1}
   \centering
   \begin{tabular}{@{}lllll@{}}
    \\
        \toprule
        &\multicolumn{1}{l}{\textbf{Acronym}} 
        & \multicolumn{1}{c}{\textbf{Extended}}
        & \multicolumn{1}{c}{\textbf{Source}}
        & \multicolumn{1}{c}{}\\
        
        \midrule
        \multirow{8}{*}{
           \rotatebox[origin=c]{90}{
               \textbf{Faithfulness}
               }
           } 
       & F.E & Faithfulness &\citet{Alvarez-Melis2018TowardsNetworks} \\
       & P.F & Pixel Flipping &\citet{Bach2015OnPropagation}\\
       & Ins.&  Insertion AUC &\citet{Petsiuk2018RISE:Models}\\
       & Del.& Deletion AUC &\citet{Petsiuk2018RISE:Models}\\
       & Ins-Del.& Insertion-Deletion AUC &\citet{Englebert2022Poly-CAM:Networks}\\
       & IROF & IROF &\citet{Rieger2020IROF:Methods}\\
       & Suff.&Sufficiency & \citet{Dasgupta2022FrameworkExplanations}\\
       & Inf. & Infidelity& \citet{Yeh2019OnExplanations}\\
           
       \midrule 
   
        \multirow{5}{*}{
           \rotatebox[origin=c]{90}{
               \textbf{Robustness}
               }
           } 

        & L. Est. & Local Lipschitz Estimate&\citet{Alvarez-Melis2018OnMethods}\\
        & M. Sens. &Max Sensitivity&\citet{Yeh2019OnExplanations}\\
        & A. Sens. &Avg. Sensitivity&\citet{Yeh2019OnExplanations}\\
        & RIS. & Relative Input Stability &\citet{Agarwal2022RethinkingExplanations}\\
        & ROS. &  Relative Output Stability &\citet{Agarwal2022RethinkingExplanations}\\
        
       \midrule 
        \vspace{4pt}  
        \multirow{2}{*}{
           \rotatebox[origin=c]{90}{
               \textbf{Com.}
               }
           } 
        & CP. & Complexity &\citet{Bhatt2020EvaluatingExplanations}\\
        & SP. & Sparseness &\citet{Chalasani2018ConciseTraining}\\
        
       \midrule 
   
        \multirow{4}{*}{
           \rotatebox[origin=c]{90}{
               \textbf{Loc.}
               }
           } 

        & A.L. & Attribution Localization &\citet{Kohlbrenner2019TowardsLRP}\\
        & T-K.L.&Top-K Intersection &\citet{Theiner2021InterpretableGeolocation}\\
        & RR-A. & Relevance Rank Accuracy &\citet{Arras2020GroundCLEVR-XAI}\\
        & RM-A.&Relevance Mass Accuracy &\citet{Arras2020GroundCLEVR-XAI}\\
        
       \midrule 
       
        & R.T. & Running Time\\

       \bottomrule
   \end{tabular}
\end{table}

% nomenclature --------------

\section{Extended Quantitative Evaluation}
\label{sec:supp-quantitative}

We verified the effectiveness of our technique across a large set of metrics, datasets and benchmarking models to assess different explanatory qualities.
Firstly, we quantified the \textit{faithfulness} aspects by computing the \textit{insertion} and \textit{deletion} AUC(s) \cite{Petsiuk2018RISE:Models} on a large poolset. We then compare the results with respect to the \textit{Faithfulness Estimate} \cite{Alvarez-Melis2018TowardsNetworks}, \textit{Pixel Flipping} \cite{Bach2015OnPropagation}, \textit{IROF} \cite{Rieger2020IROF:Methods}, \textit{Sufficiency} \cite{Dasgupta2022FrameworkExplanations} and \textit{Infidelity} \cite{Yeh2019OnExplanations}. The \textit{robustness} has been evaluated according to the \textit{Local Lipschitz Estimate} \cite{Alvarez-Melis2018OnMethods}, \textit{Max-Sensitivity}, \textit{Avg-Sensitivity} \cite{Yeh2019OnExplanations}, \textit{Relative Input Stability (RIS)}, Relative Output Stability (ROS) \cite{Agarwal2022RethinkingExplanations}. The complexity characteristic has been measured according to the \textit{Sparseness} \cite{Chalasani2018ConciseTraining} and \textit{Complexity} criteria \cite{Bhatt2020EvaluatingExplanations}. 
\textit{Insertion} and \textit{deletion} metrics have been computed using the IROF library \cite{Rieger2020IROF:Methods}, while the other metrics using the \textit{Quantus} framework v0.4.4. \cite{Hedstrom2022Quantus:Beyond}.
\textbf{Notably.} \textit{F.E} has been adopted for a more fair comparison as it is known to exhibit rank-order conflicts \cite{Rong2022AMethods, Hedstrom2023TheMetaQuantus} with similar metrics (\eg \textit{P.F}).
Due to space constraints we have attached the extended results below. The attribution baseline methods Grad-CAM, Grad-CAM++, Smooth Grad-CAM++, XGrad-CAM, Layer-CAM, Score-CAM, for Integrated Grad-CAM the code from the official repository has been adopted.

\begin{figure}
\begin{center}
    \large
    \def\svgwidth{\columnwidth}
    %% Creator: Inkscape 1.3.2 (091e20e, 2023-11-25), www.inkscape.org
%% Accompanies image file 'qualitative-examples-0.pdf' (pdf, eps, ps)
\begingroup%
  \makeatletter%
  \providecommand\color[2][]{%
    \errmessage{(Inkscape) Color is used for the text in Inkscape, but the package 'color.sty' is not loaded}%
    \renewcommand\color[2][]{}%
  }%
  \providecommand\transparent[1]{%
    \errmessage{(Inkscape) Transparency is used (non-zero) for the text in Inkscape, but the package 'transparent.sty' is not loaded}%
    \renewcommand\transparent[1]{}%
  }%
  \providecommand\rotatebox[2]{#2}%
  \newcommand*\fsize{\dimexpr\f@size pt\relax}%
  \newcommand*\lineheight[1]{\fontsize{\fsize}{#1\fsize}\selectfont}%
  \ifx\svgwidth\undefined%
    \setlength{\unitlength}{904.96850201bp}%
    \ifx\svgscale\undefined%
      \relax%
    \else%
      \setlength{\unitlength}{\unitlength * \real{\svgscale}}%
    \fi%
  \else%
    \setlength{\unitlength}{\svgwidth}%
  \fi%
  \global\let\svgwidth\undefined%
  \global\let\svgscale\undefined%
  \makeatother%
  \begin{picture}(1,0.29542264)%
    \lineheight{1}%
    \setlength\tabcolsep{0pt}%
    \put(0,0){\includegraphics[width=\unitlength,page=1]{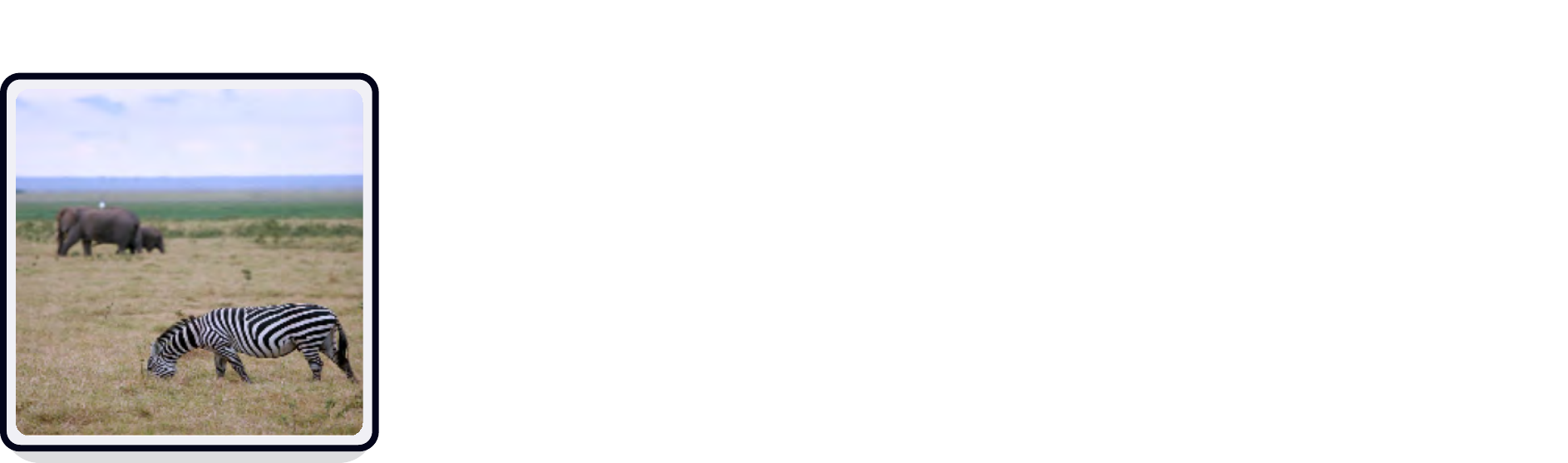}}%
    \put(0.12057529,0.27169296){\color[rgb]{0.01176471,0.01960784,0.10196078}\makebox(0,0)[t]{\lineheight{1.25}\smash{\begin{tabular}[t]{c}Image\end{tabular}}}}%
    \put(0,0){\includegraphics[width=\unitlength,page=2]{qualitative-examples-0.pdf}}%
    \put(0.37334343,0.27169296){\color[rgb]{0.01176471,0.01960784,0.10196078}\makebox(0,0)[t]{\lineheight{1.25}\smash{\begin{tabular}[t]{c}Seg. Mask\end{tabular}}}}%
    \put(0,0){\includegraphics[width=\unitlength,page=3]{qualitative-examples-0.pdf}}%
    \put(0.6261116,0.27169296){\color[rgb]{0.01176471,0.01960784,0.10196078}\makebox(0,0)[t]{\lineheight{1.25}\smash{\begin{tabular}[t]{c}Grad-CAM\end{tabular}}}}%
    \put(0,0){\includegraphics[width=\unitlength,page=4]{qualitative-examples-0.pdf}}%
    \put(0.87887977,0.27169296){\color[rgb]{0.01176471,0.01960784,0.10196078}\makebox(0,0)[t]{\lineheight{1.25}\smash{\begin{tabular}[t]{c}Ours\end{tabular}}}}%
  \end{picture}%
\endgroup%

    \normalsize
\end{center}
\caption{Example of generated binary segmentation mask for the label "zebra" from the MS-COCO dataset against Grad-CAM (baseline) and Expected Grad-CAM (our). Our technique retains and consistently exhibits low-noise properties on separate datasets.}
\label{fig:segm-mask-bin}
\end{figure}

In \Cref{table:ins-del-ext} are shown the extended \textit{faithfulness} results across the three benchmarking models, while in \Cref{table:localization-agg} are presented the findings of the localization metrics. In \Cref{fig:segm-mask-bin} is shown an example of a generated binary segmentation masks. As we employed a binary mask, the results of RM-A \cite{Arras2020GroundCLEVR-XAI} are comparable to A.L \cite{Kohlbrenner2019TowardsLRP} which we propose in \cref{table:localization-rm-a}. The relative robustness (RIS/ROS) results are tabulated in \cref{table:robustness-ris-ros}. Ultimately, the infidelity aspect has also been additionally verified on the CIFAR-10 and its results showed in \cref{table:cifar-infidelity}.

\enlrearged{
  % % supp-robust --------------
  \begin{table}
\caption{Faithfulness Metrics: Insertion and Deletion \cite{Petsiuk2018RISE:Models} AUCs computed on $5000$ samples of \textit{ILSVRC2012} \cite{Russakovsky2014ImageNetChallenge} on VGG16 \cite{Simonyan2014VeryRecognition}, ResNet-50 \cite{He2015DeepRecognition} and AlexNet \cite{Krizhevsky2012ImageNetNetworks}. \textbf{Boldface values indicate best scores}.}
\label{table:ins-del-ext}
% \small

   \centering
   % \begin{tabular}{llccc}
   \begin{tabular}{@{}lccccccccc@{}}
    \\
        \toprule
        
        &\multicolumn{3}{c}{VGG16} 
        &\multicolumn{3}{c}{ResNet-50}
        & \multicolumn{3}{c}{AlexNet}\\
        
        \cmidrule(r){2-4}
        \cmidrule(r){5-7}
        \cmidrule(r){8-10}
        
        \multicolumn{1}{l}{\textbf{Method}} 
        & \multicolumn{1}{c}{\textbf{$\uparrow$ Ins.}}
        & \multicolumn{1}{c}{\textbf{$\downarrow$ Del}}
        & \multicolumn{1}{c}{\textbf{$\uparrow$ Ins-Del.}}
        
        & \multicolumn{1}{c}{\textbf{$\uparrow$ Ins.}}
        & \multicolumn{1}{c}{\textbf{$\downarrow$ Del}}
        & \multicolumn{1}{c}{\textbf{$\uparrow$ Ins-Del.}}
        
        & \multicolumn{1}{c}{\textbf{$\uparrow$ Ins.}}
        & \multicolumn{1}{c}{\textbf{$\downarrow$ Del}}
        & \multicolumn{1}{c}{\textbf{$\uparrow$ Ins-Del.}}\\

        \midrule
           \tins Grad-CAM &$ 0.60 $&$ 0.09 $&$ 0.51 $ & 0.86 & 0.21 & 0.65 & 0.50 & 0.17 & 0.32\\
           \tins Grad-CAM++ &$ 0.58 $&$ 0.10 $&$ 0.49 $ & 0.84 & 0.21 & 0.63 & 0.48 & 0.18 & 0.30\\
           \tins Smooth Grad-CAM++ &$ 0.44 $&$ 0.17 $&$ 0.27 $ & 0.74 & 0.30 & 0.45 & 0.36 & 0.28 & 0.09\\
           \tins Integrated Grad-CAM &$ 0.61$ &$ 0.09 $&$ 0.52 $ & 0.86 & 0.21 & 0.65 & 0.51 & 0.17 & 0.34\\
           \midrule
           \tins HiRes-CAM  &$ 0.57 $&$ 0.10 $&$ 0.47 $ & 0.86 & 0.21 & 0.65 & 0.49 & 0.18 & 0.32\\
           \tins XGrad-CAM &$ \underline{0.62} $&$ 0.09 $&$ \underline{0.53}$ & 0.86 & \underline{0.2097} & 0.65 & 0.51 & 0.16 & 0.35\\
           
           \tins LayerCAM &$ 0.57 $&$ 0.10 $&$ 0.47 $ & 0.83 & 0.22 & 0.61 & 0.47 & 0.19 & 0.28\\
           \tins Score-CAM &$ 0.56 $&$ 0.11 $&$ 0.46 $ & 0.83 & 0.23 & 0.60 & 0.51 & 0.1522 & \underline{0.3554}\\
           \tins Ablation-CAM &$ 0.57 $&$ 0.10 $&$ 0.48 $ & 0.85 & 0.21 & 0.64 & 0.50 & 0.17 & 0.33\\
           \midrule 
           \tins  Expected Grad-CAM &$ \textbf{0.65} $&$ \textbf{0.09} $&$ \textbf{0.56} $ & \textbf{0.87} & \textbf{0.2093} & \textbf{0.66} & \textbf{0.52} & \underline{0.1569} & \textbf{0.3556}\\
       \bottomrule
   \end{tabular}
\normalsize
\end{table}

  % % supp-robust --------------

  % % localization --------------
  \begin{table}
\caption{Localization Metrics: scores computed on 500 samples on the MS-COCO \cite{Lin2014MicrosoftContext} dataset on VGG16 \cite{Simonyan2014VeryRecognition}, ResNet-50 \cite{He2015DeepRecognition} and AlexNet \cite{Krizhevsky2012ImageNetNetworks}. Computed on labels "zebra" and "stop sign". \textbf{Boldface values indicate best scores}. }
\label{table:localization-agg}
\small
   \centering
   \begin{tabular}{@{}llccccccccc@{}}
    \\
        \toprule
        
        &\multicolumn{3}{c}{VGG16} 
        &\multicolumn{3}{c}{ResNet-50}
        & \multicolumn{3}{c}{AlexNet}\\
        
        \cmidrule(r){2-4}
        \cmidrule(r){5-7}
        \cmidrule(r){8-10}
        
        \multicolumn{1}{l}{\textbf{Method}} 
        & \multicolumn{1}{c}{\textbf{$\uparrow$ A.L.}}
        & \multicolumn{1}{c}{\textbf{$\uparrow$ T-K.I.}} 
        & \multicolumn{1}{c}{\textbf{$\uparrow$ RR-A}}
        
        & \multicolumn{1}{c}{\textbf{$\uparrow$ A.L.}}
        & \multicolumn{1}{c}{\textbf{$\uparrow$ T-K.I.}} 
        & \multicolumn{1}{c}{\textbf{$\uparrow$ RR-A}}
        
        & \multicolumn{1}{c}{\textbf{$\uparrow$ A.L.}}
        & \multicolumn{1}{c}{\textbf{$\uparrow$ T-K.I.}} 
        & \multicolumn{1}{c}{\textbf{$\uparrow$ RR-A}}\\
        
        \midrule
           \tins Grad-CAM & 0.11 & 0.24 & 0.24 & 0.09 & 0.11 & 0.12 &0.09 & 0.07 & 0.1 & \\
           \tins Grad-CAM++ & 0.13 & 0.30 & 0.29 & \textbf{0.106} & 0.11 & 0.128 & 0.08 & 0.03 & 0.07\\
           \tins Smooth Grad-CAM++ & 0.10 & 0.18 & 0.19 & 0.07 & 0.11 & 0.12 & 0.08 & 0.03 & 0.06 \\
           \tins Integrated Grad-CAM & 0.12 & 0.34 & 0.31 & 0.097 & \underline{0.119} & 0.13 & 0.08 & 0.07 & 0.1\\
           \tins HiRes-CAM  & 0.11 & 0.22 & 0.23 & 0.097 & 0.11 & 0.12 & 0.08 & 0.04 & 0.08\\
           \tins XGrad-CAM & 0.11 & 0.24 & 0.24 & 0.09 & 0.11 & 0.12 &  0.08 & 0.05 & 0.08\\
           \midrule
           
           \tins LayerCAM & 0.11 & 0.25 & 0.24 & 0.08 & 0.1 & 0.11 & 0.07 & 0.02 & 0.06\\
           \tins Score-CAM & 0.12 & 0.25 & 0.23 & 0.09 & 0.118 & \underline{0.132} & \underline{0.109} & \underline{0.17} & \underline{0.15}\\
           \tins Ablation-CAM & \underline{0.15} & \underline{0.36} & \underline{0.33} &  0.09 & 0.11 & 0.12 & 0.106 & 0.15 & 0.14\\
           \midrule 
           
           \tins  Expected Grad-CAM & \textbf{0.18} & \textbf{0.42} & \textbf{0.36} &  \underline{0.104} & \textbf{0.18} & \textbf{0.17} & \textbf{0.13} & \textbf{0.23} & \textbf{0.18}\\
       \bottomrule
   \end{tabular}
\normalsize
\end{table}
  % % localization --------------

  \begin{table}
   \centering
    \begin{minipage}{.46\textwidth}
\caption{Localization Metrics: Rank Mass Accuracy \cite{Arras2020GroundCLEVR-XAI} computed on 500 samples on the MS-COCO \cite{Lin2014MicrosoftContext} dataset on VGG16 \cite{Simonyan2014VeryRecognition}, ResNet-50 \cite{He2015DeepRecognition} and AlexNet \cite{Krizhevsky2012ImageNetNetworks}. Computed on labels "zebra" and "stop sign". \textbf{Boldface values indicate best scores}. }
\label{table:localization-rm-a}
\footnotesize
   % \begin{tabular}{llccc}
   \begin{tabular}{@{}lccc@{}}
    \\
        \toprule
        
        &\multicolumn{1}{c}{VGG16} 
        &\multicolumn{1}{c}{ResNet-50}
        & \multicolumn{1}{c}{AlexNet}\\
        
        \cmidrule(r){2-2}
        \cmidrule(r){3-3}
        \cmidrule(r){4-4}
        
        \multicolumn{1}{l}{\textbf{Method}} 
        & \multicolumn{1}{c}{\textbf{$\uparrow$ RM-A}}
        
        & \multicolumn{1}{c}{\textbf{$\uparrow$ RM-A}}
        
        & \multicolumn{1}{c}{\textbf{$\uparrow$ RM-A}}\\
        
        \midrule
           \tins Grad-CAM & 0.11 & 0.09 & 0.09 \\
           \tins Grad-CAM++ & 0.13 & \textbf{0.11} & 0.08\\
           \tins Smooth Grad-CAM++ & 0.10 & 0.07 & 0.08 \\
           \tins Integrated Grad-CAM & 0.12 & 0.10 & 0.08 \\
           \tins HiRes-CAM  & 0.11 & 0.10 & 0.08 \\
           \tins XGrad-CAM & 0.11 & 0.09 & 0.08\\
           \midrule
           
           \tins LayerCAM & 0.11 & 0.08 & 0.07 \\
           \tins Score-CAM & 0.12 & 0.09 & \underline{0.11} \\
           \tins Ablation-CAM & \underline{0.15} & 0.09 & 0.11\\
           \midrule 
           
           \tins  Expected Grad-CAM & \textbf{0.18} & \underline{0.11} & \textbf{0.13}\\
       \bottomrule
   \end{tabular}
\end{minipage}
\hfill
\begin{minipage}{.51\textwidth}

\caption{Robustness Metrics: RIS/ROS \cite{Agarwal2022RethinkingExplanations} computed on 500 samples on the \textit{ILSVRC2012} \cite{Russakovsky2014ImageNetChallenge} dataset on VGG-16 \cite{Simonyan2014VeryRecognition} and ResNet-50 \cite{He2015DeepRecognition}. Methods marked with a '-' have been excluded due to zero-attribution values being produced under infinitesimal perturbations. \textbf{Boldface values indicate best scores}.  }
\label{table:robustness-ris-ros}
\small
   % \begin{tabular}{llccc}
   \begin{tabular}{@{}lcccc@{}}
    \\
        \toprule
        
        &\multicolumn{2}{c}{VGG-16} 
        &\multicolumn{2}{c}{ResNet-50}\\
        
        \cmidrule(r){2-3}
        \cmidrule(r){4-5}
        
        \multicolumn{1}{l}{\textbf{Method}} 
        & \multicolumn{1}{c}{\textbf{$\downarrow$ RIS}}
        & \multicolumn{1}{c}{\textbf{$\downarrow$ ROS}}
        
        & \multicolumn{1}{c}{\textbf{$\downarrow$ RIS}}
        & \multicolumn{1}{c}{\textbf{$\downarrow$ ROS}}\\

        \midrule
        \tins Grad-CAM & 169.197 & 5527.376 & 103.162 & 1.55e+04 \\ 
\tins Grad-CAM++ & 0.045 & \underline{1.3} & 357.893 & 3130.042 \\ 
\tins Smooth Grad-CAM++ & 25.003 & 2.704 & 59.733 & 1180.478  \\ 
\tins Integrated Grad-CAM & - & - & - & -  \\ 
\tins Hi-Res CAM & - & - & - & -  \\ 
\tins XGrad-CAM & 33.872 & 2812.874 & 111.022 & 1.65e+04  \\ 
        \midrule
        \tins LayerCAM & \underline{0.023} & 33.782 & \underline{11.712} & \underline{555.22}  \\ 
\tins Score-CAM & 0.09 & 14.97 & 19.046 & 2053.248  \\ 
\tins Ablation-CAM & - & - & - & -  \\ 
        \midrule 
       \tins Expected Grad-CAM & \textbf{0.004} & \textbf{0.12} & \textbf{0.573} & \textbf{73.934} \\ 
       \bottomrule
   \end{tabular}
\normalsize
\end{minipage}

\end{table}

}

% % supp-inf + rt --------------
\begin{table}
  \small
    \begin{minipage}{.55\textwidth}
     \caption{Faithfulness Metrics: Infidelity \cite{Yeh2019OnExplanations} computed on 500 samples on the CIFAR10 \cite{Ho-Phuoc2018CIFAR10Humans} dataset on VGG-16 \cite{Simonyan2014VeryRecognition}, ResNet-50 \cite{He2015DeepRecognition} and AlexNet \cite{Krizhevsky2012ImageNetNetworks}. Samples have been upsampled to $96\times96$ and the \textit{Infidelity} metric has been computed using a perturbation patch size of $32$ instead of $56$. Due to the sample low resolution, results' values are high. For readability all values have been divided by \SI{1e+7}, \SI{1e+8} and \SI{1e+9} for VGG-16, ResNet-50 and AlexNet respectively. }
   \label{table:cifar-infidelity}
   \begin{tabular}{@{}lccc@{}}
    \\
        \toprule
        
        &\multicolumn{1}{c}{VGG16} 
        &\multicolumn{1}{c}{ResNet-50}
        & \multicolumn{1}{c}{AlexNet}\\
        
        \cmidrule(r){2-2}
        \cmidrule(r){3-3}
        \cmidrule(r){4-4}
        
        \multicolumn{1}{l}{\textbf{Method}} 
        & \multicolumn{1}{c}{\textbf{$\downarrow$ Inf.}}
        
        & \multicolumn{1}{c}{\textbf{$\downarrow$ Inf.}}
        
        & \multicolumn{1}{c}{\textbf{$\downarrow$ Inf.}}\\
        
        \midrule
        \tins Grad-CAM & 1592.0 & 94.2 & 594.6 \\ 
        \tins Grad-CAM++ & 1506.5 & 88.9 & 542.0 \\ 
        \tins Smooth Grad-CAM++ & 1673.5 & 82.9 & \underline{479.0} \\ 
        \tins Integrated Grad-CAM & 1.64e+09 & 3.79e+08 & 4.83e+08 \\ 
        \tins Hi-Res CAM & 1585.6 & 77.6 & 594.1 \\ 
        \tins XGrad-CAM & 1549.2 & 93.5 & 575.0 \\ 
        
           \midrule
           \tins LayerCAM & \underline{1457.0} & 92.5 & 555.1 \\ 
            \tins Score-CAM & 1751.7 & 157.3 & 656.6 \\ 
            \tins Ablation-CAM & 1670.2 & \underline{76.5} & 619.7 \\ 
           
           \midrule 
           \tins Expected Grad-CAM & \textbf{4.7} & \textbf{3.8} & \textbf{9.6}\\ 
           
       \bottomrule
   \end{tabular}
\end{minipage}
\hfill
\begin{minipage}{.34\textwidth}
  \normalsize
  \vspace{4.4em}
   \caption{Running time computed on 100 sequential runs on the CIFAR10 \cite{Ho-Phuoc2018CIFAR10Humans} dataset on VGG-16. Averaged values are displayed.}
   \label{table:running-times}
   \begin{tabular}{@{}lc@{}}
    \\
        \toprule

        \multicolumn{1}{l}{\textbf{Method}} 
        & \multicolumn{1}{c}{\textbf{$\downarrow$ R.T.}}\\

        \midrule
        \tins Grad-CAM & \textbf{0.006}\\ 
        \tins Grad-CAM++ & \textbf{0.006} \\ 
        \tins Smooth Grad-CAM++ & 0.121 \\ 
        \tins Integrated Grad-CAM & 0.156\\
        \tins Hi-Res CAM &\textbf{0.006}\\ 
        \tins XGrad-CAM & \textbf{0.006}\\
        
           \midrule
           \tins LayerCAM & \textbf{0.006}\\
            \tins Score-CAM & 0.261 \\
            \tins Ablation-CAM & 0.302\\
           
           \midrule 
           \tins Expected Grad-CAM & \underline{0.115}\\
           
       \bottomrule
   \end{tabular}
  \end{minipage}

\end{table}

% % supp-inf + rt --------------

% -----------------------
% -----------------

\subsection{Internal Saturation}
\label{sec:internal-saturation}

\begin{figure}
  \centering

    % --- fake entry
    \begin{subfigure}[b]{0.1\textwidth}
        \captionlistentry{}
        \label{subfig:int-sat-samples}
    \end{subfigure}
    \begin{subfigure}[b]{0.1\textwidth}
        \captionlistentry{}
      \label{subfig:int-sat-featurescaling}
    \end{subfigure}
    % --- fake entry

    % \footnotesize
    %
    \def\svgwidth{\columnwidth}
    %% Creator: Inkscape 1.3.2 (091e20e, 2023-11-25), www.inkscape.org
%% Accompanies image file 'supp-saturation-samples.pdf' (pdf, eps, ps)
\begingroup%
  \makeatletter%
  \providecommand\color[2][]{%
    \errmessage{(Inkscape) Color is used for the text in Inkscape, but the package 'color.sty' is not loaded}%
    \renewcommand\color[2][]{}%
  }%
  \providecommand\transparent[1]{%
    \errmessage{(Inkscape) Transparency is used (non-zero) for the text in Inkscape, but the package 'transparent.sty' is not loaded}%
    \renewcommand\transparent[1]{}%
  }%
  \providecommand\rotatebox[2]{#2}%
  \newcommand*\fsize{\dimexpr\f@size pt\relax}%
  \newcommand*\lineheight[1]{\fontsize{\fsize}{#1\fsize}\selectfont}%
  \ifx\svgwidth\undefined%
    \setlength{\unitlength}{447.98630121bp}%
    \ifx\svgscale\undefined%
      \relax%
    \else%
      \setlength{\unitlength}{\unitlength * \real{\svgscale}}%
    \fi%
  \else%
    \setlength{\unitlength}{\svgwidth}%
  \fi%
  \global\let\svgwidth\undefined%
  \global\let\svgscale\undefined%
  \makeatother%
  \begin{picture}(1,0.91497684)%
    \lineheight{1}%
    \setlength\tabcolsep{0pt}%
    \put(0,0){\includegraphics[width=\unitlength,page=1]{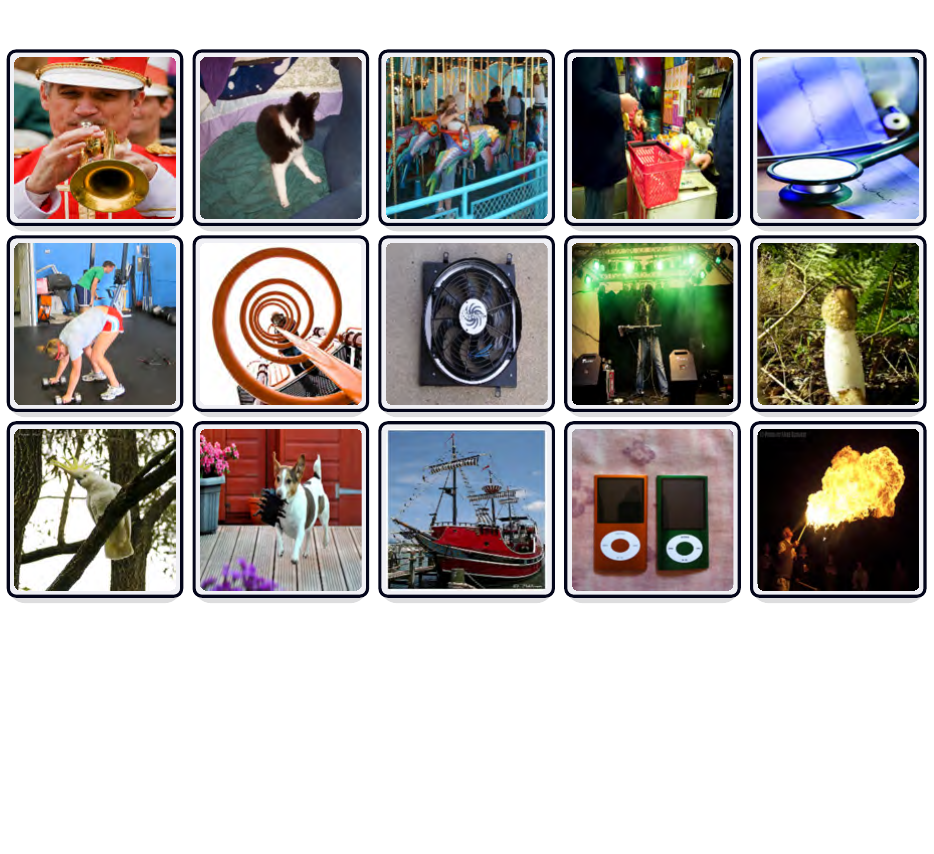}}%
    \put(0.50013907,0.88948318){\color[rgb]{0,0,0}\makebox(0,0)[t]{\lineheight{1.25}\smash{\begin{tabular}[t]{c}Samples\end{tabular}}}}%
    \put(0,0){\includegraphics[width=\unitlength,page=2]{supp-saturation-samples.pdf}}%
    \put(0.49967298,0.21852693){\color[rgb]{0,0,0}\makebox(0,0)[t]{\lineheight{1.25}\smash{\begin{tabular}[t]{c}(a)\end{tabular}}}}%
    \put(0.49967298,0.00563729){\color[rgb]{0,0,0}\makebox(0,0)[t]{\lineheight{1.25}\smash{\begin{tabular}[t]{c}(b)\end{tabular}}}}%
    \put(0,0){\includegraphics[width=\unitlength,page=3]{supp-saturation-samples.pdf}}%
    \put(0.49476469,0.10343683){\color[rgb]{0,0,0}\transparent{0.80000001}\makebox(0,0)[t]{\lineheight{1.25}\smash{\begin{tabular}[t]{c}...\end{tabular}}}}%
    \put(0,0){\includegraphics[width=\unitlength,page=4]{supp-saturation-samples.pdf}}%
  \end{picture}%
\endgroup%

    % \normalsize

  \caption{Excerpt of 25 random samples from \textit{ILSVRC2012} \cite{Russakovsky2014ImageNetChallenge} (\ref{subfig:int-sat-samples}) used to evaluate internal saturation at various points. \Cref{subfig:int-sat-featurescaling} presents a subset of samples generated through feature scaling over 25 steps.}
  \label{fig:int-saturation-samples}
\end{figure}

Following \cite{Sundararajan2016GradientsCounterfactuals} we evaluated the saturation at various points on modernly pretrained VGG-16 \cite{Simonyan2014VeryRecognition}, ResNet-50 \cite{He2015DeepRecognition} and AlexNet \cite{Krizhevsky2012ImageNetNetworks}. In \Cref{subfig:int-sat-samples} are shown the $25$ random samples utilized, alongside a selected excerpt of the samples generated using the following feature scaling procedure (\cref{subfig:int-sat-featurescaling}) for $N=25$:
$$\left\{\alpha_i \mid \alpha_i \sim U(0,1), \; i=1,2, \ldots, N\right\}$$

\begin{figure}
  \centering
    % --- fake entry
    \begin{subfigure}[b]{0.1\textwidth}
        \captionlistentry{}
        \label{subfig:saturation2-cmp-0}
    \end{subfigure}
    \begin{subfigure}[b]{0.1\textwidth}
        \captionlistentry{}
        \label{subfig:saturation2-cmp-1}
    \end{subfigure}
    % --- fake entry

    \footnotesize
    \def\svgwidth{\columnwidth}
    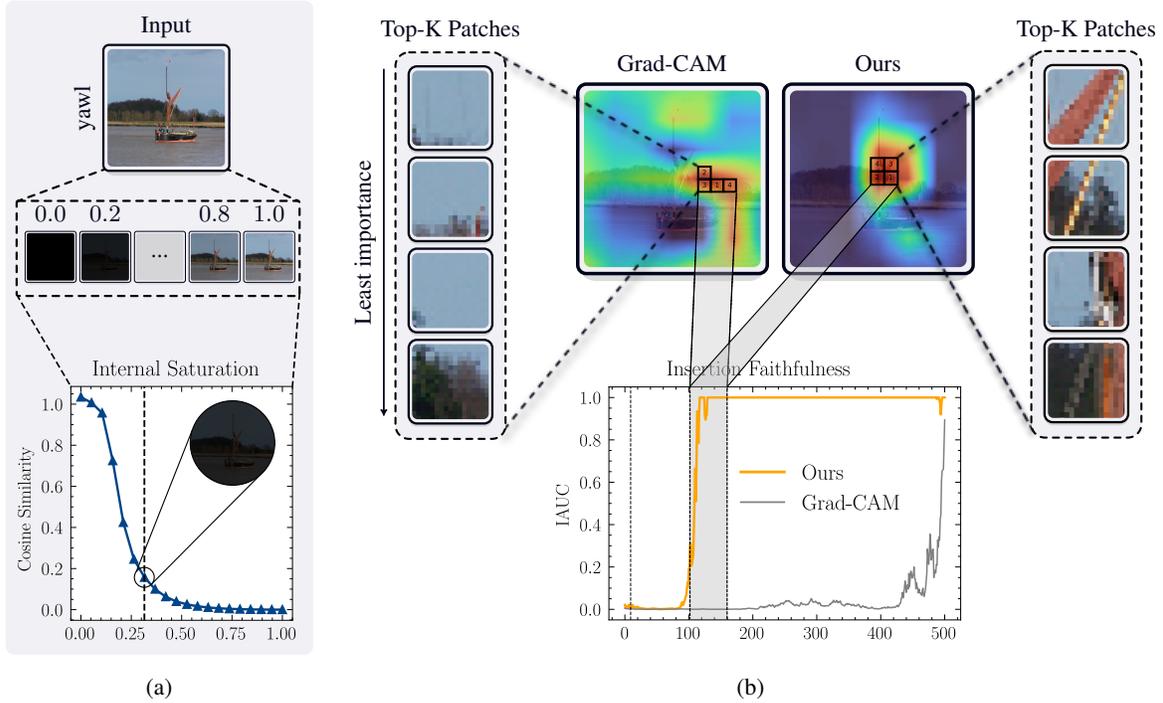

    \normalsize

    \caption{Comparison of the attribution maps under internal saturation conditions. In \Cref{subfig:saturation-cmp-0} is shown the cosine similarity of the target layer's embeddings with respect to the interpolator parameter ($\alpha$). \Cref{subfig:saturation-cmp-1} shows the attribution maps of the different methods under the saturation condition. The internal saturation condition causes the baseline method to under-represent feature importances across saturating ranges. By extracting the top-4 most important features (\cref{subfig:saturation-cmp-1}) we can observe that the baseline method fails to capture the relevant discriminative regions, which produce low insertion AUCs (\cref{subfig:saturation-cmp-1}) as deemed not important by the model.}

   \label{fig:saturation-cmp-alt}
\end{figure}

\begin{figure}
  \centering

    % --- fake entry
    \begin{subfigure}[b]{0.1\textwidth}
        \captionlistentry{}
      \label{subfig:internal-sat}
    \end{subfigure}
    \begin{subfigure}[b]{0.1\textwidth}
        \captionlistentry{}
      \label{subfig:internal-sat-mean}
    \end{subfigure}
    % --- fake entry

    \vspace*{-2em}
    \def\svgwidth{\columnwidth}
    %% Creator: Inkscape 1.3.2 (091e20e, 2023-11-25), www.inkscape.org
%% Accompanies image file 'sat-study-internal.pdf' (pdf, eps, ps)
\begingroup%
  \makeatletter%
  \providecommand\color[2][]{%
    \errmessage{(Inkscape) Color is used for the text in Inkscape, but the package 'color.sty' is not loaded}%
    \renewcommand\color[2][]{}%
  }%
  \providecommand\transparent[1]{%
    \errmessage{(Inkscape) Transparency is used (non-zero) for the text in Inkscape, but the package 'transparent.sty' is not loaded}%
    \renewcommand\transparent[1]{}%
  }%
  \providecommand\rotatebox[2]{#2}%
  \newcommand*\fsize{\dimexpr\f@size pt\relax}%
  \newcommand*\lineheight[1]{\fontsize{\fsize}{#1\fsize}\selectfont}%
  \ifx\svgwidth\undefined%
    \setlength{\unitlength}{444.19887272bp}%
    \ifx\svgscale\undefined%
      \relax%
    \else%
      \setlength{\unitlength}{\unitlength * \real{\svgscale}}%
    \fi%
  \else%
    \setlength{\unitlength}{\svgwidth}%
  \fi%
  \global\let\svgwidth\undefined%
  \global\let\svgscale\undefined%
  \makeatother%
  \begin{picture}(1,1.26675479)%
    \lineheight{1}%
    \setlength\tabcolsep{0pt}%
    \put(0,0){\includegraphics[width=\unitlength,page=1]{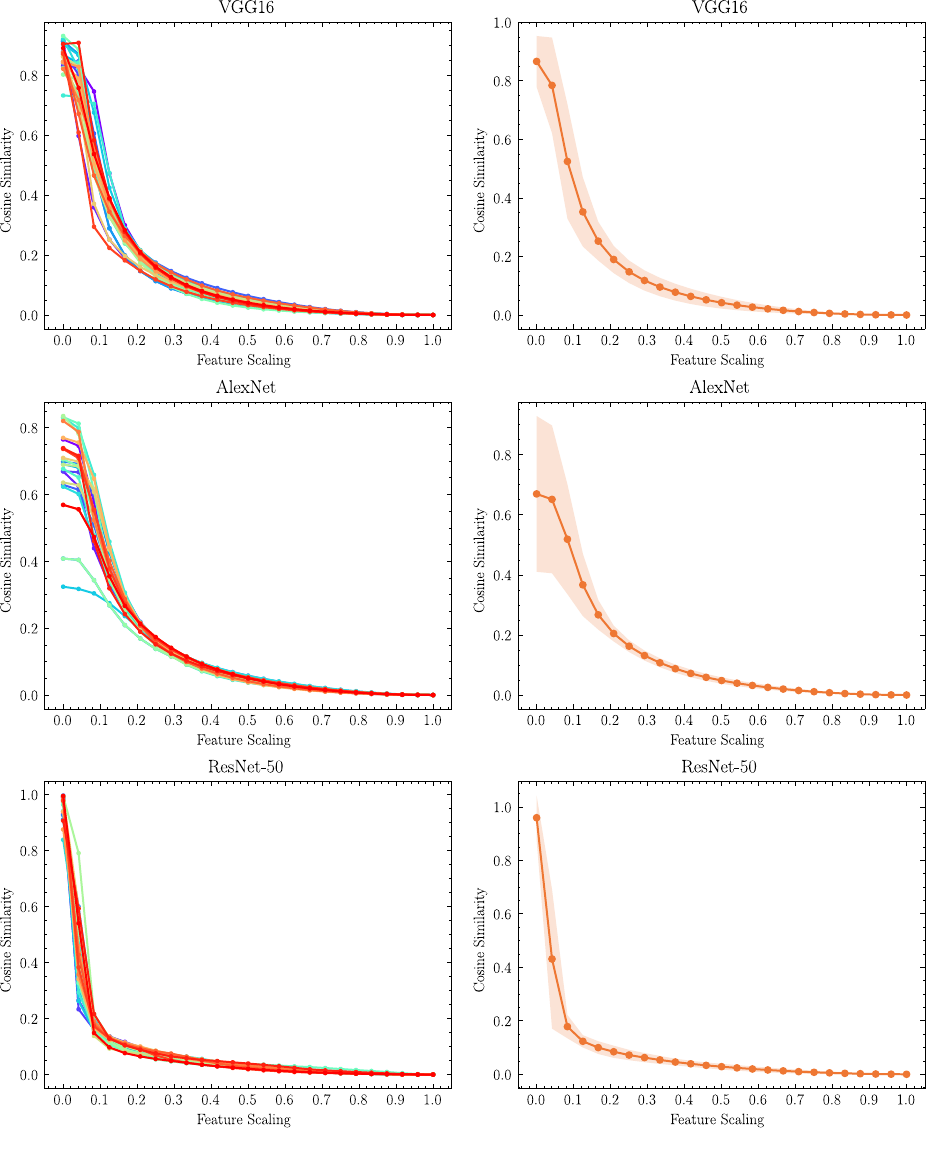}}%
    \put(0.27014885,0.00568516){\color[rgb]{0,0,0}\makebox(0,0)[t]{\lineheight{1.25}\smash{\begin{tabular}[t]{c}(a)\end{tabular}}}}%
    \put(0.78173606,0.00568516){\color[rgb]{0,0,0}\makebox(0,0)[t]{\lineheight{1.25}\smash{\begin{tabular}[t]{c}(b)\end{tabular}}}}%
  \end{picture}%
\endgroup%

  \caption{Internal saturation analysis of intermediary target layers in VGG16 \cite{Simonyan2014VeryRecognition}, AlexNet \cite{Krizhevsky2012ImageNetNetworks}, and ResNet-50 \cite{He2015DeepRecognition}. \Cref{subfig:internal-sat} presents the cosine similarity between activation vectors of CAM target filters. \Cref{subfig:internal-sat-mean} depicts the mean values with error bars indicating 2 standard deviations. For VGG16 and AlexNet, the final feature layer is used, while for ResNet-50, the $layer4$ is selected.}
\label{fig:int-saturation-intermediary}

\end{figure}

\begin{figure}
  \centering

   % --- fake entry
    \begin{subfigure}[b]{0.1\textwidth}
        \captionlistentry{}
      \label{subfig:output-sat}
    \end{subfigure}
    \begin{subfigure}[b]{0.1\textwidth}
        \captionlistentry{}
      \label{subfig:output-sat-mean}
    \end{subfigure}
    % --- fake entry

    \vspace*{-2em}
    \def\svgwidth{\columnwidth}
    %% Creator: Inkscape 1.3.2 (091e20e, 2023-11-25), www.inkscape.org
%% Accompanies image file 'sat-study-output.pdf' (pdf, eps, ps)
\begingroup%
  \makeatletter%
  \providecommand\color[2][]{%
    \errmessage{(Inkscape) Color is used for the text in Inkscape, but the package 'color.sty' is not loaded}%
    \renewcommand\color[2][]{}%
  }%
  \providecommand\transparent[1]{%
    \errmessage{(Inkscape) Transparency is used (non-zero) for the text in Inkscape, but the package 'transparent.sty' is not loaded}%
    \renewcommand\transparent[1]{}%
  }%
  \providecommand\rotatebox[2]{#2}%
  \newcommand*\fsize{\dimexpr\f@size pt\relax}%
  \newcommand*\lineheight[1]{\fontsize{\fsize}{#1\fsize}\selectfont}%
  \ifx\svgwidth\undefined%
    \setlength{\unitlength}{444.20030008bp}%
    \ifx\svgscale\undefined%
      \relax%
    \else%
      \setlength{\unitlength}{\unitlength * \real{\svgscale}}%
    \fi%
  \else%
    \setlength{\unitlength}{\svgwidth}%
  \fi%
  \global\let\svgwidth\undefined%
  \global\let\svgscale\undefined%
  \makeatother%
  \begin{picture}(1,1.32044714)%
    \lineheight{1}%
    \setlength\tabcolsep{0pt}%
    \put(0,0){\includegraphics[width=\unitlength,page=1]{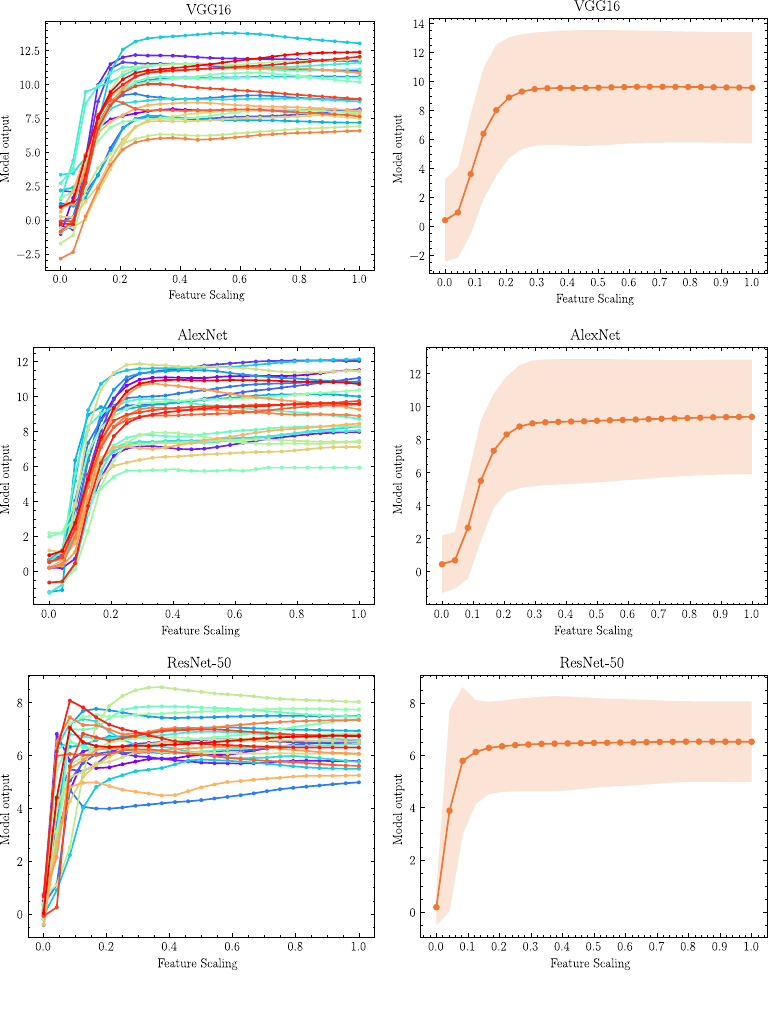}}%
    \put(0.26491314,0.00568524){\color[rgb]{0,0,0}\makebox(0,0)[t]{\lineheight{1.25}\smash{\begin{tabular}[t]{c}(a)\end{tabular}}}}%
    \put(0.7770997,0.00568524){\color[rgb]{0,0,0}\makebox(0,0)[t]{\lineheight{1.25}\smash{\begin{tabular}[t]{c}(b)\end{tabular}}}}%
  \end{picture}%
\endgroup%

\caption{Output saturation analysis in VGG16 \cite{Simonyan2014VeryRecognition}, AlexNet \cite{Krizhevsky2012ImageNetNetworks}, and ResNet-50 \cite{He2015DeepRecognition} (\cref{subfig:output-sat}). \Cref{subfig:output-sat} displays the softmax scores for the top label, while \Cref{subfig:output-sat-mean} depicts the mean values with error bars indicating 2 standard deviations.}

  \label{fig:int-saturation-output}
\end{figure}

\Cref{fig:int-saturation-output,fig:int-saturation-intermediary} shows the saturating behavior \wrt the output and intermediary layers targeted by CAM methods. Both the \textit{pre-softmax} and \textit{post-softmax} outputs quickly flatten and plateaus for very small value of the feature scaling factor $\alpha$, with the softmax outputs showing the swiftest rate of change and abruptly converge to saturation (\cref{fig:int-saturation-output}). When selecting an arbitrary intermediary layer (\ie the one targeted by the analyzed CAM methods) the saturation phenomena is still present but offset due to the reduced path (depth) (\cref{fig:saturation-cmp-alt}).
As $\alpha$ increases, the cosine similarity of the target layer's embeddings quickly flattens (\Cref{subfig:saturation2-cmp-0}), leading to an underestimation of feature attributions. This results in sparse, uninformative, and ill-formed explanations (\Cref{subfig:saturation2-cmp-1}). This is evident when inspecting the top-k most important patches according to the generated attribution maps, which focus on background areas rather than the target class (\textit{yawl}). Consequently, when these patches are inserted, they produce low model confidence (Insertion IAUC) (\Cref{subfig:saturation2-cmp-1}). Conversely, our method focuses on salient discriminative areas of the image that characterize the target label (\ie \textit{yawl}) and highly activate the neural network, demonstrating high fidelity to the model's inner workings, robustness to internal saturation, and high localization by focusing only on the most important regions.

% ++++++++++++++++++++++++++++++++++++
% ++++++++++++++++++++++++++++++++++++

%-------------------------
%-------------------------

% page style ----------------------
\newpage
\pagestyle{plain}
% page style ----------------------

\def\w{.49}
\def\linewidth{\textwidth}

%-------------------------
% %-------------------------

\newcommand{\h}[0]{0.7}
%-------------------------
%-------------------------

% page style ----------------------
\pagestyle{empty}
% page style ----------------------

\clearpage

% page style ----------------------
\pagestyle{plain}
% page style ----------------------

\section{Qualitative Evaluation}
\label{sec:supp-qualitative}

Next we provide the extended version of all the figures \ie including all the comparative baseline methods and some additional examples.

%-------------------------
\begin{figure}
  \centering

    % \footnotesize
    %
    \def\svgwidth{\columnwidth}
    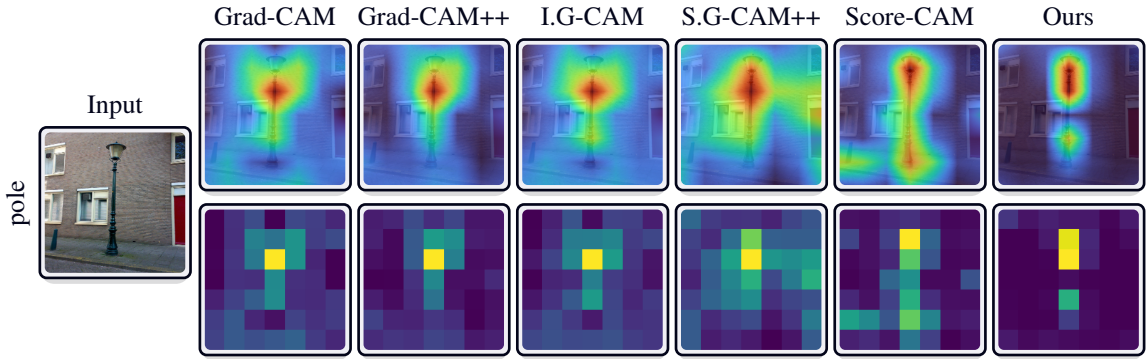

    % \normalsize

  \caption{Gradients are noisy. A comparison of gradient-based CAM methods under optimal conditions shows that even recent methods exhibit high sensitivity.}
  \label{fig:qualitative-examples-2}
\end{figure}
%-------------------------

% -----------------------
\begin{figure}
  \centering
  \includegraphics[width=0.85\columnwidth]{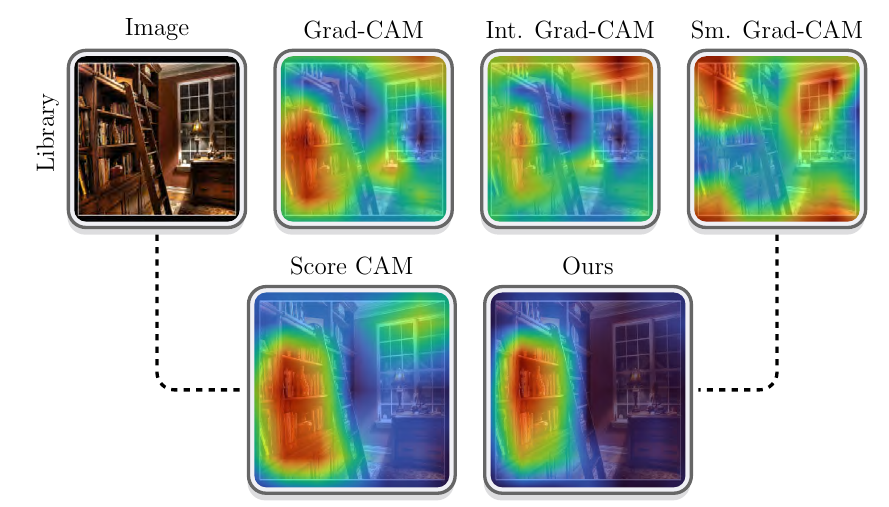}
  \vspace{1em}
  
   \caption{Comparison of gradient-based and non-gradient-based CAM methods. The top row illustrates the noisiness and tendency to produce ill-formed explanations in gradient-based methods, including recent approaches \cite{Sattarzadeh2021IntegratedScoring}. \textit{Score-CAM} \cite{Wang2019Score-CAM:Networks} addresses this issue by eliminating the use of gradients. Our method demonstrates the ability to generate sharper and more stable explanations consistently, even with the use of gradients.}
  \label{fig:noisy-gradients}
\end{figure}
% -----------------

% qualitative extended comparison page --------------
\enlrearged{
  %-------------------------
  \begin{figure}
    \begin{center}
        \scriptsize
    \def\svgwidth{\columnwidth}
    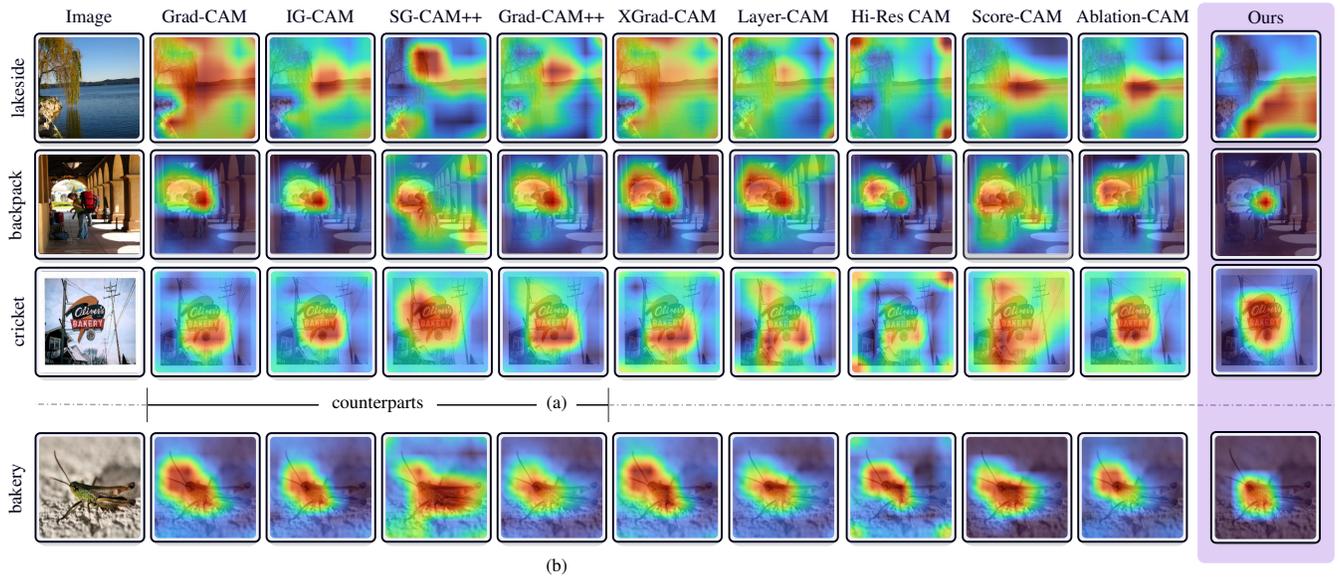

        \normalsize
    \end{center}

      % --- fake entry
      \begin{subfigure}[b]{0.1\textwidth}
          \captionlistentry{}
          \label{subfig:qual-eval-ext-0}
      \end{subfigure}
      \begin{subfigure}[b]{0.1\textwidth}
          \captionlistentry{}
          \label{subfig:qual-eval-ext-1}
      \end{subfigure}
      % --- fake entry

    \caption{Comparison of the attribution maps for various methods under normal (\ref{subfig:qual-eval-ext-1}) and internal saturation condition (\ref{subfig:qual-eval-ext-0}). Extended version containing all baseline attribution methods.}
    \label{fig:qualitative-comparison-extended}
  \end{figure}
  %-------------------------

  %-------------------------
  \begin{figure}
    \centering
    \includegraphics[width=\columnwidth]{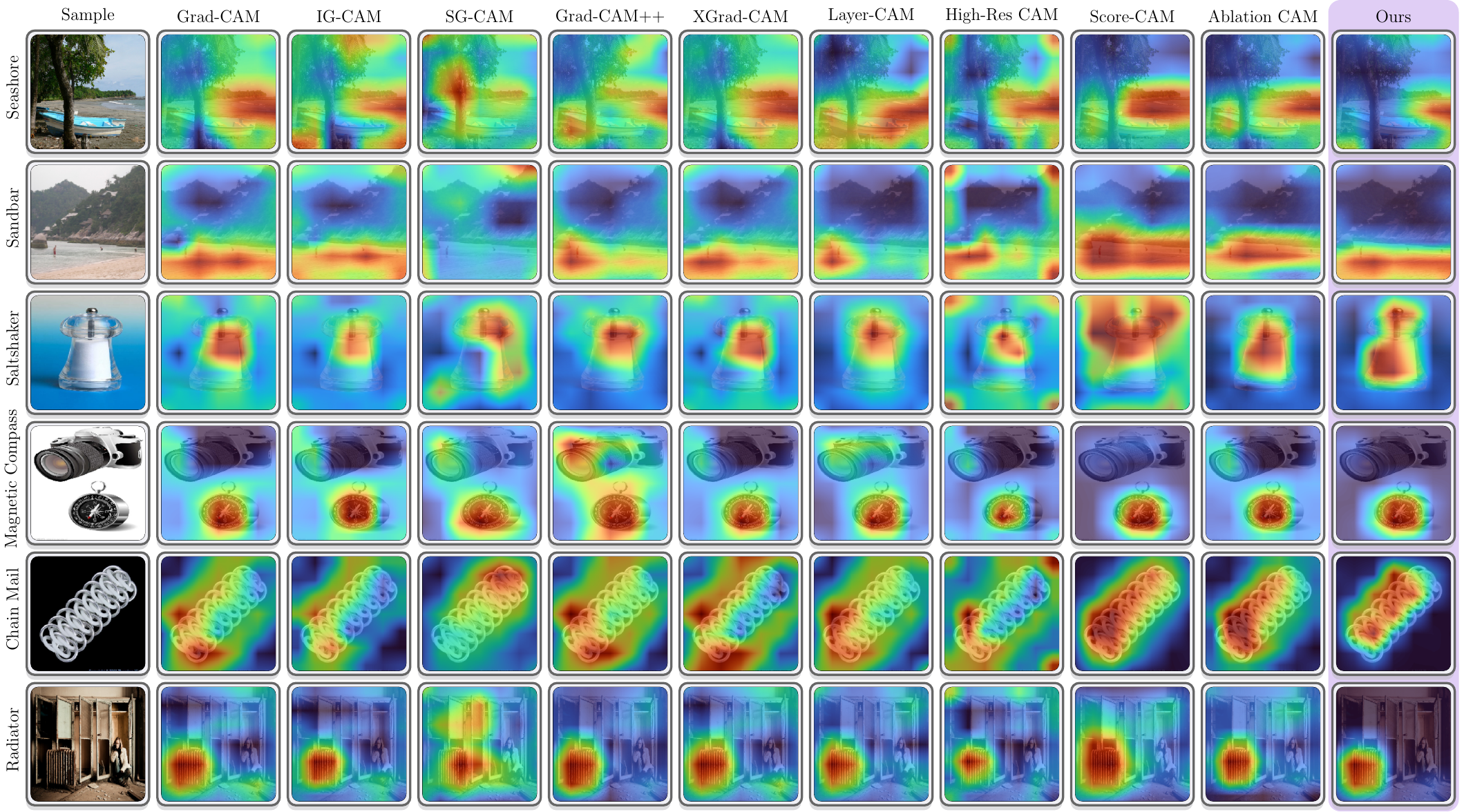}
      
    \caption{Comparison of saliency maps between our method and all the baseline methods on the \textit{ILSVRC2012} \cite{Russakovsky2014ImageNetChallenge}.}
    \label{fig:qualitative-more}
  \end{figure}
  %-------------------------
}

%%%%%%%%%%%%%%%%%%%%%%%%%%%%%%%%%%%%%%%%%%%%%%%%%%%%%%%%%%%%

\end{document}